\documentclass[10pt,journal,compsoc]{IEEEtran}
\usepackage{amsmath,amsfonts}
\usepackage{algorithmic}
\usepackage{algorithm}
\usepackage{array}
\usepackage[caption=false,font=normalsize,labelfont=sf,textfont=sf]{subfig}
\usepackage{textcomp}
\usepackage{stfloats}
\usepackage{url}
\usepackage{verbatim}
\usepackage{graphicx}
\usepackage{booktabs}

\usepackage{multirow}
\usepackage{multicol}
\usepackage[table]{xcolor}
\definecolor{methodgray}{gray}{0.94}
\usepackage{hyperref}
\hypersetup{
    pagebackref=true,
	colorlinks=true,
	linkcolor=red,
	filecolor=blue,      
	urlcolor=red,
	citecolor=green,
}
\usepackage{cite}
\newcommand{\red}[1]{\textcolor{red}{#1}}

\newcommand{\green}[1]{\textcolor[RGB]{96,177,87}{#1}}

\newlength\savewidth

\begin{document}

\title{UniVAD v2: Unified Visual Anomaly Detection via Support-Conditioned Boundary Construction}

\author{Zhaopeng Gu, Bingke Zhu, Zhaowen Li, Guibo Zhu, Yingying Chen,  Ming Tang,  \IEEEmembership{Member,~IEEE}, \\ Peng Su, Jinqiao Wang, \IEEEmembership{Member,~IEEE}



\thanks{Zhaopeng Gu and Ming Tang are with the Foundation Model Research Center, Institute of Automation, Chinese Academy of Sciences, Beijing 100190, China, and also with the School of Artificial Intelligence, University of Chinese Academy of Sciences, Beijing 100049, China (e-mail: guzhaopeng2023@ia.ac.cn; tangm@nlpr.ia.ac.cn).}
\thanks{Bingke Zhu and Yingying Chen are at the Foundation Model Research Center, Institute of Automation, Chinese Academy of Sciences, Beijing 100190, China (e-mail: bingke.zhu@nlpr.ia.ac.cn; yingying.chen@nlpr.ia.ac.cn).}
\thanks{Zhaowen Li and Peng Su are at the Yinwang Intelligent Technology Co. Ltd., China (e-mail: lizhaowen3@huawei.com; p.su@huawei.com).}
\thanks{Guibo Zhu is with the Foundation Model Research Center, Institute of Automation, Chinese Academy of Sciences, Beijing 100190, China, also with the School of Artificial Intelligence, University of Chinese Academy of Sciences, Beijing 100049, China, and also with the Shanghai Artificial Intelligence Laboratory, Shanghai 200232, China (e-mail: gbzhu@nlpr.ia.ac.cn).}
\thanks{Jinqiao Wang is with the Foundation Model Research Center, Institute of Automation, Chinese Academy of Sciences, Beijing 100190, China, also with the School of Artificial Intelligence, University of Chinese Academy of Sciences, Beijing 100049, China, also with the Wuhan AI Research, Wuhan 430073, China, also with the Peng Cheng Laboratory, Shenzhen 518066, China, and also with the Guangdong Provincial Key Laboratory of Intellectual Property \& Big Data, Guangdong Polytechnic Normal University, Guangzhou 510665, China (e-mail: jqwang@nlpr.ia.ac.cn).}
}

\markboth{IEEE Transactions on Pattern Analysis and Machine Intelligence}{Gu \MakeLowercase{\textit{et al.}}: UniVAD v2: Unified Visual Anomaly Detection via Support-Conditioned Boundary Construction}


\IEEEtitleabstractindextext{
\begin{abstract}
Unified visual anomaly detection seeks to train a single detector that can be deployed across categories, domains, and application scenarios. In the few-shot transfer regime, the key challenge is to estimate an episode-specific boundary for an unseen target category from a small support set. Existing approaches mainly infer this boundary from normal-side evidence and provide limited abnormal-side evidence for deployment-specific tolerance. Within the normal side, they often struggle to jointly capture local correspondences and global support-query relations, making their boundaries less reliable for unseen anomalies. To address these issues, we propose UniVAD v2, a two-sided support-conditioned boundary construction framework for unified visual anomaly detection. Built on the component-patch divide-and-conquer framework of UniVAD, UniVAD v2 strengthens the normal side with an \textbf{O}ptimal \textbf{T}ransport-based \textbf{R}elational \textbf{M}odeling module~(OTRM), which complements retrieval with support-query matching through transport-style allocation, and an \textbf{A}daptive \textbf{C}oordination mechanism for \textbf{R}etrieval and \textbf{R}elational \textbf{M}odeling~(ACRRM), which estimates episode-conditioned reliabilities to fuse the two sources of evidence. On the abnormal side, a \textbf{F}ew-Shot \textbf{A}bnormal \textbf{R}eference module~(FAR) converts optional abnormal references into rejection-side evidence for boundary adjustment. Experiments on six datasets spanning industrial, logical, and medical anomaly detection demonstrate strong cross-domain generalization. Under the 1N-shot protocol, UniVAD v2 improves the mean image-level AUC over UniVAD from 83.0\% to 84.5\%, and further reaches 85.7\% in the 1N+1A-shot setting. On the MVTec-AD Severity Split~(MVTec-AD-SS), UniVAD v2 achieves 96.2\% image-level AUC and 96.9\% pixel-level AUC, showing that abnormal references enable controllable boundary customization without retraining.
\end{abstract}


}

\maketitle
\IEEEdisplaynontitleabstractindextext

\section{Introduction}

\begin{figure}[t]
    \centering
    \includegraphics[width=\linewidth]{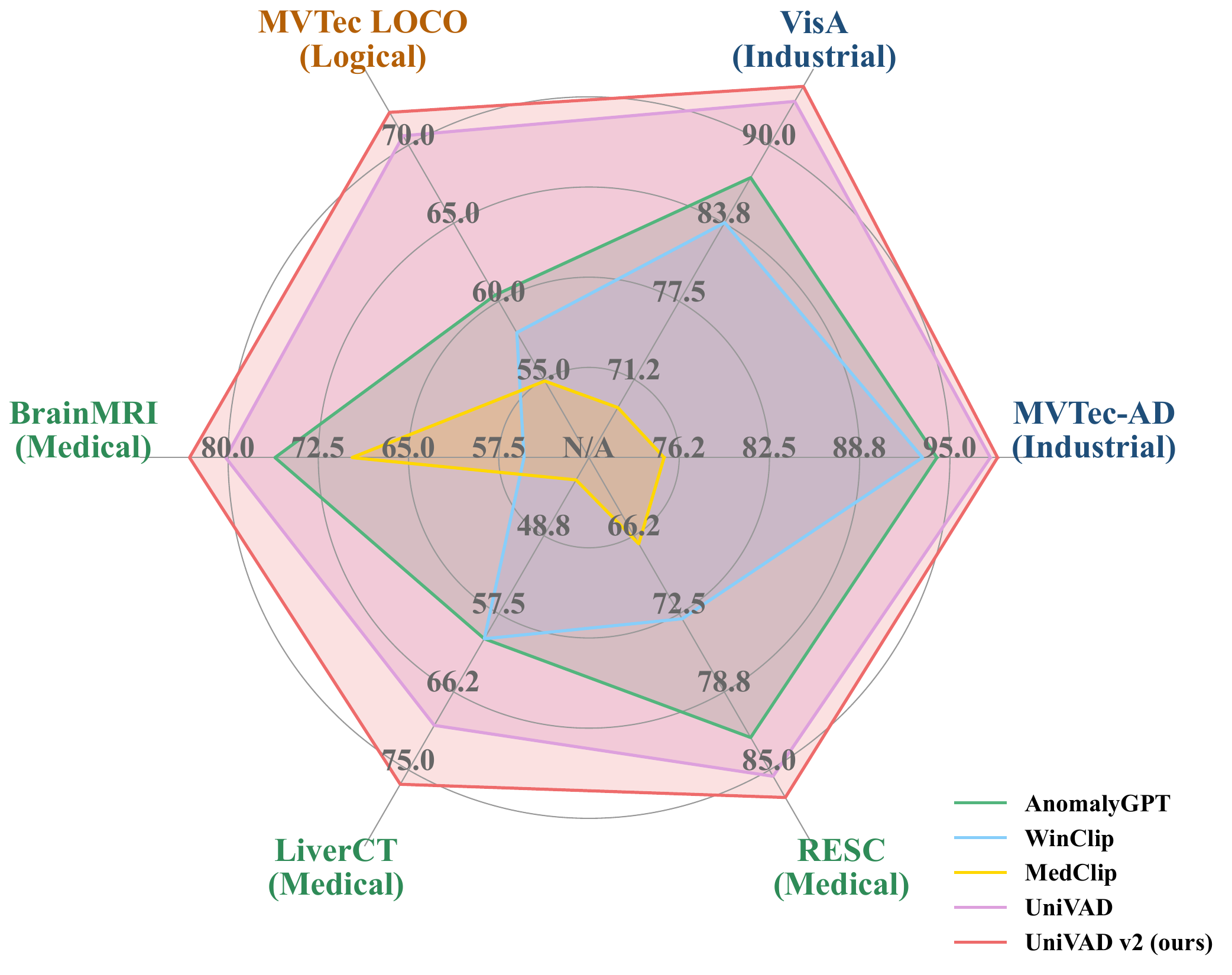}
    \caption{Overall comparison of anomaly detection methods on six datasets spanning industrial, logical, and medical scenarios. The radar plot summarizes image-level AUC on each benchmark, where a larger enclosed region indicates a stronger and more balanced cross-domain performance profile. UniVAD v2 forms the outer frontier on all axes, highlighting its consistent advantage across heterogeneous settings.}
    \label{fig:radar}
\end{figure}

\IEEEPARstart{V}{isual} anomaly detection is a fundamental topic in computer vision and artificial intelligence. Its core goal is to identify samples that deviate from normal patterns, supporting applications in intelligent manufacturing~\cite{roth2022towards, gu2024anomalygpt, gu2026anomalymoe, gu2024filo, gu2026filo++, zhu2024pixel, qian2026quality}, medical image analysis~\cite{bao2024bmad, huang2024adapting, su2021few}, and safety monitoring~\cite{georgescu2021anomaly, rong2024dam, wu2024open}. Traditional anomaly detection methods~\cite{defard2021padim, deng2022anomaly, gudovskiy2022cflow} usually follow a ``one-class-one-model'' paradigm, where a separate detector is trained for each category or scenario. This design becomes increasingly restrictive when the target category, acquisition condition, or application domain changes. Unified anomaly detection methods~\cite{gu2025univad, guo2025dinomaly, wei2025uninet, fu2025reason, willes2022bayesian} therefore aim to build a single detector that can be reused across heterogeneous settings. In practical few-shot deployment, the detector is often applied to an unseen target category with only a small number of reference samples. The central problem is thus to construct a reliable decision boundary for the current support episode.

\begin{figure*}[t]
    \centering
    \includegraphics[width=\linewidth]{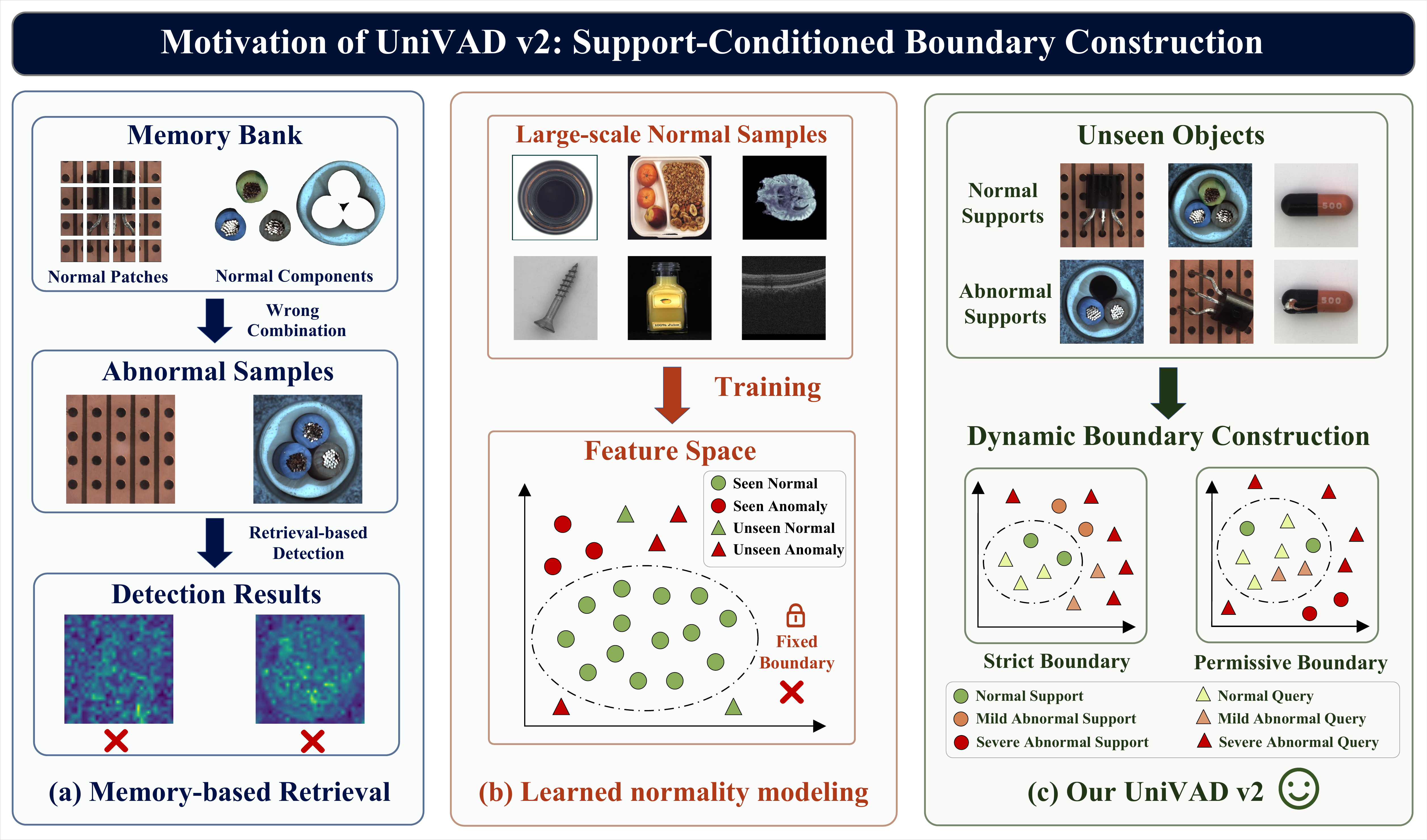}
    \caption{Comparison of boundary-evidence mechanisms in unified anomaly detection. (a) Memory-based retrieval builds memory banks from normal patches or components, but locally plausible matches may still form a wrong global combination and lead to missed anomalies. (b) Learned normality modeling estimates an effective boundary from large-scale normal samples, but the boundary is fixed after training and transfers weakly to unseen objects with different tolerance requirements. (c) UniVAD v2 constructs the boundary from normal supports and optional abnormal supports at inference time. Normal supports provide evidence for the acceptable side, while abnormal supports provide residual rejection-side evidence; different abnormal-support severities can induce stricter or more permissive effective boundaries for unseen objects.}
    \label{fig:motivation}
\end{figure*}

As shown in Fig.~\ref{fig:motivation}, existing anomaly detection methods mainly infer this boundary from normal-side evidence. Memory-based retrieval methods~\cite{roth2022towards, gu2025univad, cai2026raid, gu2024filo} provide stable support matching, but their evidence is assembled from independent patch or component correspondences and may miss logical anomalies caused by wrong combination of normal components. Learned normality modeling methods~\cite{guo2025dinomaly, wei2025uninet, gu2026anomalymoe} encode broader context, but their effective scoring boundaries are typically fixed after training and transfer weakly to unseen objects or changing tolerance requirements. Thus, retrieval and learned normality modeling provide complementary but incomplete normal-side evidence. Normal-side evidence alone also leaves the rejection side of the boundary implicit. In real inspection systems, the desired tolerance may vary across deployments, where mild defects can be acceptable in one setting but rejected in another. Abnormal references naturally provide evidence about which deviations should be rejected. Recent methods such as NAGL~\cite{wang2025normal} use normal-abnormal references to improve generalist anomaly detection, but abnormal samples are mainly treated as additional guidance for strengthening anomaly scores. How to combine abnormal-reference evidence with normal-side support evidence for support-conditioned boundary adjustment remains underexplored.

To address this problem, we propose UniVAD v2, a support-conditioned boundary construction framework for unified visual anomaly detection. On the normal side, UniVAD v2 combines retrieval with an Optimal Transport-based Relational Modeling module~(OTRM), which introduces global support-query matching through transport-style allocation. An Adaptive Coordination mechanism for Retrieval and Relational Modeling~(ACRRM) predicts episode-conditioned fusion weights to coordinate local retrieval evidence and global relational evidence. On the abnormal side, the Few-Shot Abnormal Reference module~(FAR) converts optional abnormal references into residual rejection-side evidence, allowing the effective boundary to change with the references provided at inference time.

We evaluate UniVAD v2 on six datasets spanning industrial inspection~\cite{bergmann2019mvtec, zou2022spot}, medical imaging~\cite{bao2024bmad}, and logical anomaly detection~\cite{bergmann2022beyond}. The main comparison fixes the selected normal supports for all methods, so the results measure how effectively different detectors construct the boundary from the same normal-support evidence on unseen target categories. Under this fixed 1N-shot protocol, UniVAD v2 improves the six-dataset mean image-level AUC from 83.0\% to 84.5\% over UniVAD. With one abnormal reference, UniVAD v2 further reaches 85.7\%. We then use controlled reference variations to analyze the behavior of the constructed boundary. On MVTec-AD-SS, UniVAD v2 achieves 96.2\% image-level AUC and 96.9\% pixel-level AUC, demonstrating support-conditioned boundary customization without retraining.

Our contributions are summarized as follows:
\begin{enumerate}
  \item We formulate unified visual anomaly detection as support-conditioned boundary construction, where the inference-time decision boundary for unseen categories is estimated from the evidence available in the current support episode.
  \item We strengthen normal-side boundary construction with ACRRM and OTRM, where retrieval provides stable local support evidence, OT-based relational modeling provides global support-query matching evidence, and adaptive coordination selects their reliability under each episode.
  \item We extend support-conditioned boundary construction to the abnormal side with FAR, which converts optional abnormal references into residual rejection-side evidence and enables support-conditioned boundary customization without retraining.
  \item Experiments on six datasets and MVTec-AD-SS verify UniVAD v2's cross-domain generalization, component complementarity, support scalability, and dynamic boundary adjustment under different abnormal-reference conditions.
\end{enumerate}

\section{Related Work}

\subsection{Anomaly Detection}
The growing need for anomaly detection systems that can transfer across products, imaging conditions, and application domains makes conventional ``one-class-one-model'' methods inadequate. As a result, anomaly detection is moving toward unified methods that can handle multi-category data across domains with a single model. From the perspective of support-conditioned boundary construction, current unified anomaly detection methods mainly differ in how they obtain boundary evidence for a target episode.

Memory-based retrieval is a common way to construct normal-side evidence from reference samples~\cite{cohen2020sub, defard2021padim, shi2020query}. PaDiM~\cite{defard2021padim} is an early representative of this line of work. PatchCore~\cite{roth2022towards} substantially reduces memory size through coreset sampling, achieving accurate local anomaly localization through nearest-neighbor search. To address more complex scenarios, recent studies adopt stronger retrieval mechanisms. RegAD~\cite{huang2022registration} uses STN-based~\cite{jaderberg2015spatial} image registration before retrieval to improve feature correspondence, while UniVAD~\cite{gu2025univad} and RAID~\cite{cai2026raid} construct hierarchical memory banks to perform matching at multiple levels, including global, component, and patch levels. These methods are highly accurate for local structural defects such as scratches and stains, but they still rely heavily on independent patch or component correspondences. Consequently, when anomalies arise from incorrect combinations of otherwise normal features, local matching may miss them because plausible matches still exist in the normal memory bank.

Learned normality modeling provides another source of boundary evidence by learning reconstruction mappings, scoring functions, or decision surfaces from training data~\cite{madan2023self, guo2025dinomaly, wei2025uninet, gu2026anomalymoe, zhu2024adformer}. Dinomaly~\cite{guo2025dinomaly} proposes a Transformer-based reconstruction framework for multi-class settings. UniNet~\cite{wei2025uninet} combines contrastive learning with domain-related feature selection to bridge the domain gap across industrial and medical data. AnomalyMoE~\cite{gu2026anomalymoe} uses a mixture-of-experts architecture to detect anomalies at different semantic levels. These methods naturally benefit from learnable global contextual modeling, but their decision surfaces are learned from normal samples in the training categories and then kept static at inference time. Such static normal-side evidence creates two limitations. First, when the model encounters unseen objects, the learned reconstruction behavior or decision function may not match the new category, so the performance can degrade sharply. Second, because the scoring behavior is fixed after training and is optimized only from normal data, it cannot be adjusted according to the abnormal references available during deployment. 

Recent work has begun to use abnormal references at inference time. NAGL~\cite{wang2025normal} proposes a normal-abnormal-guided setting for anomaly detection and introduces Residual Mining and Anomaly Feature Learning to exploit abnormal samples as auxiliary guidance for improving detection performance. This line of work shows that abnormal references are useful, but it does not directly address how test-time normal and abnormal supports should jointly shape the boundary for an unseen target episode. UniVAD v2 focuses on this underexplored problem by treating abnormal samples as abnormal-side evidence that works with normal-side support to adjust the inference-time boundary.

Overall, existing anomaly detectors provide limited mechanisms for jointly exploiting local correspondence, global support-query relations, and controllable abnormal-side references. UniVAD v2 addresses this gap by formulating unified anomaly detection as support-conditioned boundary construction, where normal supports and abnormal references play complementary roles.

\subsection{Optimal Transport}
Optimal transport~(OT) studies the minimum-cost plan for transforming one probability distribution into another. Because it preserves the geometry of the underlying feature space, OT has attracted growing interest in deep learning theory and visual correspondence. Recent work has also revealed a close connection between the Transformer architecture and optimal transport~\cite{akbari2022deep, he2024learnable, xu2025plovad}. ICRL~\cite{wang2024transformers} shows that the softmax attention layer effectively simulates gradient descent on the dual of entropy-regularized Wasserstein-2 optimal transport, which allows provable global alignment over discrete point sets. Lotformer~\cite{shahbazi2025lotformer} further derives a doubly stochastic linear attention mechanism from the optimal transport perspective and achieves exact feature alignment in linear time through a learnable pivot measure. In anomaly detection tasks, optimal transport also shows promise. RD++~\cite{tien2023revisiting} introduces a self-supervised optimal transport loss into reverse distillation to improve the compactness of normal features. Prototype-based OT~\cite{ke2024prototype} measures the distance between test inputs and distribution prototypes for out-of-distribution detection. RoDA~\cite{liao2025robust} applies the Sinkhorn distance to domain adaptation and coarse distribution alignment in industrial inspection.

Despite these advances, existing anomaly detection studies mostly use OT as a training regularizer or a static distance. Its mass-allocation view has been less explored as an inference-time matching principle for logical and structural anomaly detection. UniVAD v2 follows this direction by using OT-based relational modeling to complement independent retrieval and to introduce global support-query matching into boundary construction.

\section{Method}
\label{sec:method}

\begin{figure*}[t]
    \centering
    \includegraphics[width=0.93\linewidth]{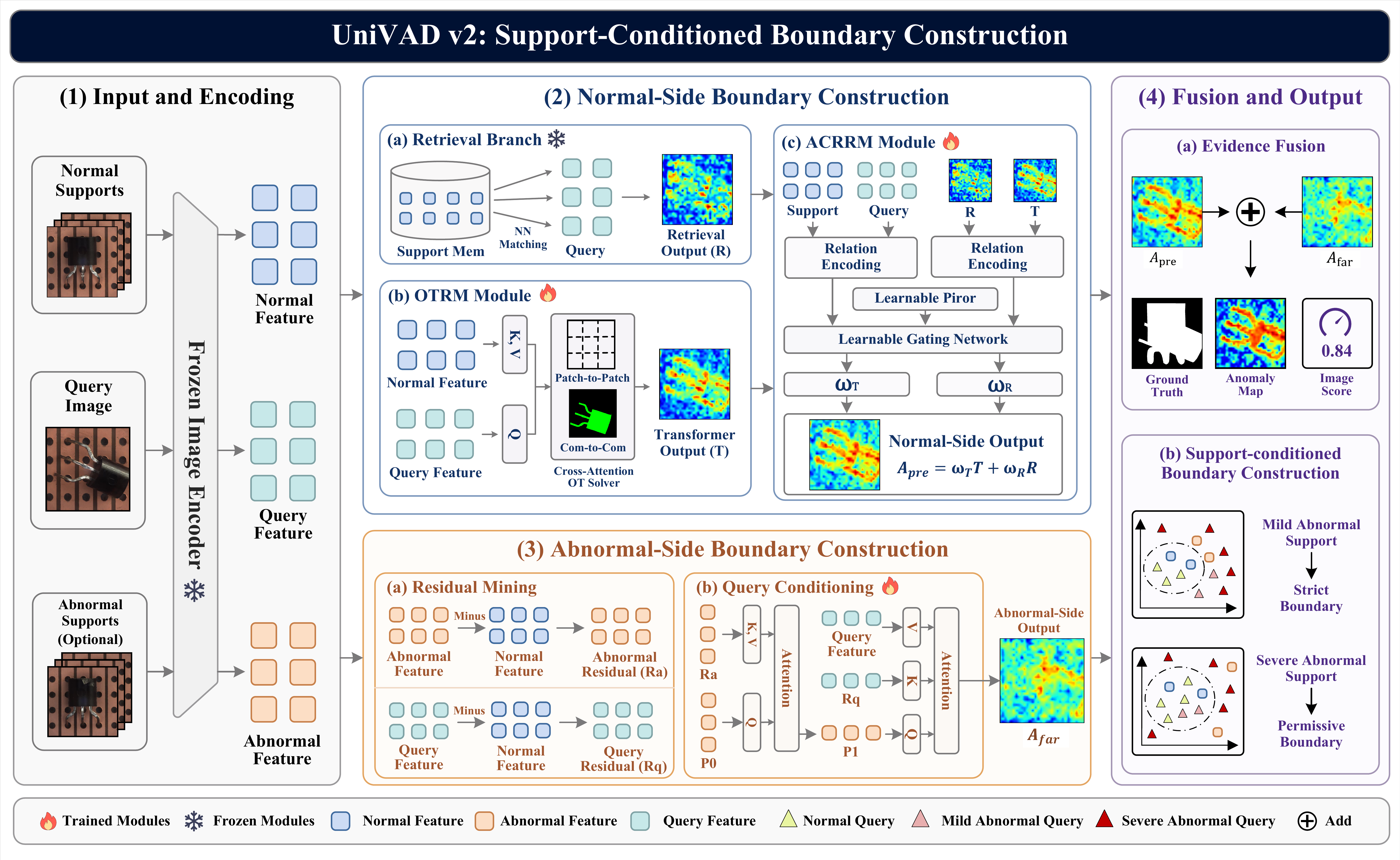}
    \caption{Overview of UniVAD v2 as support-conditioned boundary construction. In stage (1), normal supports, the query image, and optional abnormal supports are encoded by frozen image encoders into dense visual features. In stage (2), normal-side boundary construction combines a frozen retrieval branch with trainable OTRM, and ACRRM predicts the fusion weights \(\omega_R\) and \(\omega_T\) from the support-query relation and the branch responses. In stage (3), FAR converts optional abnormal supports into abnormal-side evidence through residual mining and query conditioning. In stage (4), the normal-side and abnormal-side maps are fused to obtain the final anomaly map, while the support-conditioned boundary view illustrates how mild or severe abnormal supports induce strict or permissive effective boundaries.}
    \label{fig:method_overview}
\end{figure*}

\subsection{Problem Formulation}
We consider a few-shot unified anomaly detection episode \(\mathcal{E}=(\mathcal{S}^{n},\mathcal{S}^{a},I^{q})\), where \(\mathcal{S}^{n}=\{I_i^{n}\}_{i=1}^{K_n}\) is a normal support set with \(K_n\) images, \(\mathcal{S}^{a}=\{I_j^{a}\}_{j=1}^{K_a}\) is an optional abnormal support set with \(K_a\) images, and \(I^{q}\) is a query image from a previously unseen category. When no abnormal reference is available, \(K_a=0\) and \(\mathcal{S}^{a}\) is empty. The goal is to learn an episode-conditioned predictor \(f_{\theta}\) that maps \(\mathcal{E}\) to a dense anomaly map \(A\in\mathbb{R}^{H\times W}\) and an image-level anomaly score \(s\in\mathbb{R}\), where \(H\times W\) denotes the spatial resolution of the query feature grid. Instead of relying on a static normal-side scoring behavior learned from training categories, \(f_{\theta}\) should construct decision evidence from the current support episode. The problem is therefore to estimate \(f_{\theta}\) such that the boundary can be accurately constructed from only a few references, remain transferable to unseen object categories and different domains, and be adjusted by abnormal-side evidence when abnormal references are provided.

\subsection{Framework Overview}
As illustrated in Fig.~\ref{fig:method_overview}, UniVAD v2 follows a four-stage pipeline that constructs boundary evidence from the current support episode. In the first stage, frozen image encoders extract dense visual features from the normal supports, the query image, and the optional abnormal supports. Following UniVAD, we retain the component-patch divide-and-conquer view of the task, in which patch-level correspondence captures local defects and component-level matching captures structural and logical consistency.

The second stage constructs normal-side boundary evidence from the normal supports. For each branch \(b\in\{\mathrm{loc},\mathrm{part},\mathrm{cmp}\}\), where \(b\) denotes local patch matching, component-aware patch matching, and component-level matching, respectively, the retrieval branch performs nearest-neighbor matching over the support memory bank and produces a retrieval score \(R_b\). In parallel, OTRM performs masked OT-based relational matching between support and query tokens and produces a relational modeling score \(T_b\). ACRRM then receives the support-query relation together with \(R_b\) and \(T_b\), predicts branch-wise fusion coefficients \(\omega_b^{r},\omega_b^{t}\in[0,1]\) with \(\omega_b^{r}+\omega_b^{t}=1\), and obtains the fused normal-side score as:
\begin{equation}
\tilde{A}_b=\omega_b^{r}R_b+\omega_b^{t}T_b.
\label{eq:branch_fusion}
\end{equation}
The normal-side anomaly map is defined by:
\begin{equation}
A_{\mathrm{pre}}=\frac{\tilde{A}_{\mathrm{loc}}+\tilde{A}_{\mathrm{part}}+A_{\mathrm{vl}}}{3}
+\lambda_{c}\tilde{A}_{\mathrm{cmp}},
\label{eq:unified_fusion}
\end{equation}
where \(A_{\mathrm{vl}}\) denotes the feature map obtained from image-text similarity computation, and \(\lambda_c\) controls the contribution of component-level evidence. This formulation covers images with and without meaningful component-aware evidence and therefore avoids separate hand-crafted fusion rules for different structural cases.

The third stage constructs abnormal-side boundary evidence when abnormal supports are available. FAR converts the optional abnormal supports into an abnormal-side map \(A_{\mathrm{far}}\) that indicates how strongly query regions align with reference-defined abnormal patterns. The fourth stage fuses the normal-side and abnormal-side maps, and the final prediction is:
\begin{equation}
A=A_{\mathrm{pre}}+A_{\mathrm{far}}.
\label{eq:final_map}
\end{equation}
This pipeline preserves the robustness of support-based matching, adds global relational modeling over the normal supports, and allows optional abnormal supports to adjust the effective scoring behavior at inference time.

\subsection{Adaptive Coordination Module}
\label{sec:arrcm}
The Adaptive Coordination of Retrieval and Relational Modeling module~(ACRRM) constructs normal-side boundary evidence by integrating retrieval-based and relational-modeling scores within a single episode-conditioned framework. Instead of using fixed fusion coefficients, it predicts branch-specific weights from the support-query relation and from the current outputs of the two branches. In this way, UniVAD v2 retains the stability of support-based matching while allowing the final prediction to benefit from global relational modeling when it is informative.

For each branch \(b\in\{\mathrm{loc},\mathrm{part},\mathrm{cmp}\}\), ACRRM receives a retrieval score map \(R_b\), a relational modeling score map \(T_b\), a support set \(\mathcal{S}_b\), and a query set \(\mathcal{Q}_b\). The first step is to encode the global relation between \(\mathcal{S}_b\) and \(\mathcal{Q}_b\) in a category-agnostic manner. Since support and query features form unordered sets, the relation encoder should be permutation-invariant with respect to support ordering. Following the Deep Sets~\cite{zaheer2017deep} principle, we first map each element into a latent space and then apply symmetric statistics to summarize the support set and the query set. Concretely, let \(\phi(\cdot)\) denote an element encoder. For each branch, ACRRM computes:
\begin{equation}
\begin{aligned}
\mathbf{h}^{S}_{b}&=\big[\operatorname{mean}(\phi(\mathcal{S}_b));\operatorname{max}(\phi(\mathcal{S}_b));\operatorname{std}(\phi(\mathcal{S}_b))\big],\\
\mathbf{h}^{Q}_{b}&=\big[\operatorname{mean}(\phi(\mathcal{Q}_b));\operatorname{max}(\phi(\mathcal{Q}_b));\operatorname{std}(\phi(\mathcal{Q}_b))\big],
\end{aligned}
\end{equation}
and then combines them with low-order statistics of pairwise support-query similarity to obtain the relation embedding:
\begin{equation}
\mathbf{z}_b=\rho\!\left([\mathbf{h}^{S}_{b};\mathbf{h}^{Q}_{b};\operatorname{sim}(\phi(\mathcal{S}_b),\phi(\mathcal{Q}_b))]\right),
\label{eq:arrcm_relation}
\end{equation}
where \(\rho(\cdot)\) is a nonlinear projector.

ACRRM then determines the gating value through a precision-inspired fusion rule. To motivate this design, let \(Y_b\) denote the latent anomaly evidence to be estimated in branch \(b\). In an idealized view, retrieval and relational modeling can be regarded as two noisy observations of \(Y_b\):
\begin{equation}
R_b=Y_b+\varepsilon_b^{r},\qquad T_b=Y_b+\varepsilon_b^{t}.
\label{eq:arrcm_estimators}
\end{equation}

A simplified minimum-variance fusion objective would choose the branch weights by solving:
\begin{equation}
\min_{\omega_b^{r}+\omega_b^{t}=1}\operatorname{Var}\!\left(\omega_b^{r}R_b+\omega_b^{t}T_b\right).
\label{eq:arrcm_mv}
\end{equation}

If the error covariance is further approximated by a diagonal form with scale parameters \((\sigma_b^{r})^2\) and \((\sigma_b^{t})^2\), this objective suggests inverse-variance-style weighting:
\begin{equation}
\omega_b^{r\star}=\frac{\pi_b^{r}}{\pi_b^{r}+\pi_b^{t}},\qquad \omega_b^{t\star}=\frac{\pi_b^{t}}{\pi_b^{r}+\pi_b^{t}},
\label{eq:arrcm_precision}
\end{equation}
where $\pi_b^{r}=(\sigma_b^{r})^{-2}, \pi_b^{t}=(\sigma_b^{t})^{-2}$. This idealized derivation motivates a reliability-aware gating design. In the implemented model, the precision terms are amortized as learned reliability logits conditioned on the current support-query episode, rather than estimated as explicit statistical variances. Let \(\mathbf{s}_b\) denote the statistics of \(R_b\), \(T_b\), and their discrepancy, including the mean, standard deviation, peak response, and the magnitude of disagreement. The context is instantiated as \(\mathbf{c}_b=[\mathbf{z}_b;\mathbf{s}_b]\). We further introduce a learnable embedding \(g\) to encode the branch-specific structural prior. The gate computes:
\begin{equation}
\boldsymbol{\ell}_b=\psi_b(\mathbf{c}_b)+\mathbf{p}(g),
\label{eq:arrcm_log_precision}
\end{equation}
where \(\psi_b(\cdot)\) is an MLP for episode-conditioned reliability prediction, and \(\mathbf{p}(\cdot)\) is an MLP that maps the learnable embedding \(g\) to a prior logit. To adjust the sharpness of the branch mixture, a learnable temperature \(\tau(g)\) is further introduced and the final weights are obtained by:
\begin{equation}
\left[\omega_b^{r},\omega_b^{t}\right]=\mathrm{Softmax}\!\left(\frac{\boldsymbol{\ell}_b}{\tau(g)}\right).
\label{eq:arrcm}
\end{equation}

This gating formulation implements a reliability-aware mixture of the retrieval and relational branches. The softmax operation normalizes the learned reliability logits into branch weights, and the logits are conditioned on both the support-query relation and the response statistics of the two branches. Therefore, ACRRM can assign larger weight to retrieval when local correspondence provides more discriminative evidence, and emphasize relational modeling when global support-query structure is more informative. Since the gate is computed from episode-level relations and branch responses rather than category identities, ACRRM remains category-agnostic and can be applied to unseen object categories.

\subsection{Optimal Transport-Based Relational Modeling}
\label{sec:otrm}
OTRM is responsible for strengthening the relational part of normal-side boundary construction. Its purpose is to complement isolated nearest-neighbor retrieval with a learnable global matching mechanism grounded in optimal transport theory~\cite{villani2009optimal}. Let the support and query token sets induce two empirical measures:
\begin{equation}
\mu=\sum_{u=1}^{U}a_u\,\delta_{\mathbf{s}_u}, \qquad \nu=\sum_{p=1}^{P}b_p\,\delta_{\mathbf{q}_p},
\end{equation}
where \(\delta_{\mathbf{x}}\) is the Dirac measure, \(\mathbf{a}=\{a_u\}_{u=1}^{U}\) denotes the marginal masses of the support measure \(\mu\), and \(\mathbf{b}=\{b_p\}_{p=1}^{P}\) denotes the marginal masses of the query measure \(\nu\). For consistency with the query-to-support allocation matrix used by OTRM, the coupling set is defined as:
\begin{equation}
\mathcal{U}(\mathbf{b},\mathbf{a})=
\left\{
\Pi\in\mathbb{R}_{+}^{P\times U}
\mid
\Pi\mathbf{1}_{U}=\mathbf{b},\;
\Pi^{\top}\mathbf{1}_{P}=\mathbf{a}
\right\}.
\end{equation}
Each entry \(\Pi_{p,u}\) represents the transport mass assigned from query token \(\mathbf{q}_p\) to support token \(\mathbf{s}_u\). Entropic OT seeks a transport plan \(\Pi\) that minimizes
\begin{equation}
\Pi^{\star}=\arg\min_{\Pi\in\mathcal{U}(\mathbf{b},\mathbf{a})}\langle \Pi,C\rangle+\varepsilon H(\Pi),
\label{eq:ot_objective}
\end{equation}
where \(C\in\mathbb{R}^{P\times U}\) is the query-to-support transport cost matrix and \(H(\Pi)\) is the entropy regularizer. This formulation provides a principled description of global matching under mass conservation and competition among support candidates in the support set.

Recent theory~\cite{daneshmand2024provable, wang2024transformers} shows that Transformer~\cite{vaswani2017attention} attention is closely connected to the optimization of entropic optimal transport. In particular, prior studies~\cite{daneshmand2024provable} relate softmax attention to gradient-based updates on the dual of entropy-regularized Wasserstein transport. This connection motivates OTRM as an OT-based relational matching module. In OTRM, query tokens allocate normalized mass over admissible support candidates, and structural masks define the support-query pairs that participate in the transport-style matching process.

Concretely, given support tokens \(X^{s}\) and query tokens \(X^{q}\), OTRM first projects them into latent attention spaces \(\mathbf{Q}^{\mathrm{att}}, \mathbf{K}^{\mathrm{att}}, \mathbf{V}^{\mathrm{att}}\) and constructs the following relaxed allocation matrix:
\begin{equation}
\tilde{\Pi}=\mathrm{Softmax}\!\left(\frac{\mathbf{Q}^{\mathrm{att}}{\mathbf{K}^{\mathrm{att}}}^{\top}}{\sqrt{d}}+M\right),
\label{eq:ot_attention}
\end{equation}
where \(M\) encodes structural admissibility. If a support-query pair is structurally invalid, its entry is suppressed so that \(\tilde{\Pi}_{p,u}=0\) after normalization. The resulting matrix is a masked one-sided transport relaxation, where each row describes how the mass of a query token is distributed over admissible normal-support candidates. The attended representation \(\tilde{\Pi} \cdot \mathbf{V}^{\mathrm{att}}\) summarizes the support evidence allocated to each query token, and a prediction head maps this globally matched representation to anomaly scores. Therefore, the relational modeling score is produced by a trainable OT-based inference network whose attention pattern is shaped by competition among admissible support candidates, mass allocation, and structural constraints.

This principle is instantiated at three levels. In the local branch, every query patch is matched against the whole support set, which gives OTRM a holistic view unavailable to independent nearest-neighbor retrieval. In the part branch, the attention mask \(M\) enforces component-consistent matching, so the relaxed allocation is restricted to support tokens compatible with the refined component partition. In the component branch, support and query components are treated as structured sets of higher-level descriptors and are matched through the same OT-based relational modeling principle. Because all three branches share the same global matching philosophy, OTRM can model both local defects and inter-component logical anomalies within a unified learnable formulation.

\subsection{Few-Shot Abnormal Reference Module}
\label{sec:far}
Most anomaly detectors estimate the data distribution from only the normal side of the decision boundary. This design is restrictive in industrial inspection, where the acceptable tolerance can vary across production requirements. Mild defects may be considered unacceptable in a strict quality-control setting but acceptable when the operational goal is to reject only severe defects. A detector whose boundary is determined only by normal supports has limited ability to express such deployment-specific tolerance.

FAR is designed to construct abnormal-side boundary evidence rather than to use anomalous samples for incremental performance gain. It converts abnormal references into residual rejection-side evidence and combines them with normal supports to obtain a support-conditioned anomaly score. The normal supports anchor the acceptable side, while the abnormal supports specify which residual deviations should be rejected.

FAR follows a residual anomaly modeling strategy in which anomaly evidence is extracted by subtracting normal correspondences and then distilled into a compact set of learnable proxies. Let \(K_n\) and \(K_a\) denote the numbers of normal and abnormal support images, \(P=H\times W\) denote the number of spatial tokens in each image, and \(d\) denote the feature dimension. We denote the normal support token tensor by \(S^{n}\in\mathbb{R}^{K_n\times P\times d}\), the abnormal support token tensor by \(S^{a}\in\mathbb{R}^{K_a\times P\times d}\), and the query token tensor by \(X^{q}\in\mathbb{R}^{1\times P\times d}\), where the singleton first dimension indicates that each episode contains one query image. FAR begins by constructing residual features. Let \(\mathbf{a}_{v}\) be the \(v\)-th flattened abnormal-support token and \(\mathbf{q}_{p}\) be the token at the \(p\)-th query location. For each token, we remove the closest normal-support explanation:
\begin{equation}
\mathbf{r}^{a}_{v}=\mathbf{a}_{v}-\operatorname{NN}(\mathbf{a}_{v};S^{n}), \qquad \mathbf{r}^{q}_{p}=\mathbf{q}_{p}-\operatorname{NN}(\mathbf{q}_{p};S^{n}),
\label{eq:far_residual}
\end{equation}
where \(\operatorname{NN}(\cdot;S^{n})\) denotes nearest-neighbor retrieval in the normal support set. This subtraction suppresses object identity and preserves the residual deviation from normality. This residual view is essential for boundary customization because abnormal supports are not used as raw memories of particular defect instances, but are converted into deviations relative to the current normal support manifold. FAR therefore constrains the boundary with anomaly evidence that is both reference-specific and category-agnostic.

The first stage of FAR is residual mining. A set of learnable anomaly proxies \(\mathbf{P}^{0}=\{\mathbf{p}^{0}_{m}\}_{m=1}^{M}\) attends to the abnormal support set. In the implementation, each proxy performs cross-attention over abnormal support tokens, where the abnormal residuals serve as key and values. This design lets the proxies decide where to look using the actual abnormal context while aggregating the residual signal that distinguishes abnormal regions from the normal support manifold. A self-attention refinement block is then applied to the proxy set, yielding residual proxies \(\mathbf{P}^{1}\):
\begin{equation}
\mathbf{P}^{1}=\operatorname{Attention}\!\left(\textbf{Q}=\mathbf{P}^{0},\textbf{K}=R^{a},\textbf{V}=R^{a};M^{a}\right),
\label{eq:far_rm}
\end{equation}
where \(R^{a}\) denotes the abnormal residual tokens and \(M^{a}\) is the residual mining mask.

The second stage is query conditioning, which adapts the mined abnormal proxies to the current query image. \(\mathbf{P}^{1}\) are used as queries to attend to the residual query tokens, while the original query tokens provide the values. This is again followed by self-attention refinement:
\begin{equation}
\mathbf{P}^{2}=\operatorname{Attention}\!\left(\textbf{Q}=\mathbf{P}^{1},\textbf{K}=R^{q},\textbf{V}=X^{q}\right),
\label{eq:far_afl}
\end{equation}
where \(R^{q}\) denotes the query residual tokens. This step transfers the anomaly patterns mined from the abnormal support set to the current query and adapts them to the query-specific context. Thus, FAR does not merely compare the query against a fixed abnormal memory. It first extracts anomaly prototypes and then reconditions these prototypes on the residual structure of the current query.

The final FAR score is obtained by measuring the affinity between each query token and the learned anomaly proxies:
\begin{equation}
A_{\mathrm{far}}(p)=\frac{1}{M}\sum_{m=1}^{M}\cos\!\left(\mathbf{q}_{p},\mathbf{p}^{2}_{m}\right).
\label{eq:far_score}
\end{equation}
This score is complementary to OTRM. OTRM evaluates how well the query can be explained by the normal support distribution, whereas FAR evaluates how strongly the query aligns with residual patterns distilled from real anomalies. Since the final anomaly map in Eq.~\eqref{eq:final_map} increases with abnormal evidence and the ranking loss encourages anomalous patches to receive higher scores, a larger \(A_{\mathrm{far}}\) makes a query region more likely to be rejected. The combination turns the detector into a bounded contrastive model shaped by both normal-side and abnormal-side few-shot references. Because the residual proxies are mined from the abnormal supports provided in the current episode, the resulting scoring behavior is reference-conditioned. Mild abnormal references contain subtle residual patterns, so they can raise the scores of mild deviations and induce a stricter effective boundary. Severe abnormal references emphasize stronger residual patterns; mild deviations then match the abnormal-side proxies less strongly and become more acceptable under a tolerance-driven setting, producing a more permissive effective boundary that focuses on pronounced defects. When abnormal supports are not available, FAR is disabled and the model naturally reduces to the normal-only setting governed by the normal-side score.

\subsection{Training Objective}
\label{sec:training}
Let \(Y\in\{0,1\}^{H\times W}\) be the query patch label map. UniVAD v2 is trained with a pairwise ranking objective that directly encourages anomalous query patches to receive higher scores than normal query patches:
\begin{equation}
\mathcal{L}_{\mathrm{rank}}
=
\frac{1}{|\Omega|}
\sum_{(i,j)\in\Omega}
\max\!\left(0,\gamma-A_{i}^{+}+A_{j}^{-}\right),
\label{eq:rank_loss}
\end{equation}
where \(\Omega\) is a set of sampled anomalous-normal patch pairs, \(A_{i}^{+}\) and \(A_{j}^{-}\) are anomaly scores from positive and negative query patches, and \(\gamma\) is the ranking margin. During inference, the patch-level anomaly map is upsampled to the input resolution, and the image-level anomaly score is obtained from the peak response of the dense map.

\begin{table*}[t]
\centering
\caption{Comparison with existing methods on six datasets. Image-level AUC (Img.) and pixel-level AUC (Pixel) are reported for each dataset together with their six-dataset mean (Mean) (\%). For the normal-support comparison, all methods are evaluated with the same selected normal supports. UniVAD v2 (1N-shot) provides the fixed-support fair comparison, while UniVAD v2 (1N+1A-shot) reports the reference-assisted setting with one additional abnormal reference. The best and second-best results are highlighted in bold and underlined, respectively.}
\label{tab:sota_main}
\resizebox{\textwidth}{!}{
\begin{tabular}{lcccccccccccccc}
\toprule
 \multirow{2}{*}{Method} & \multicolumn{2}{c}{MVTec-AD} & \multicolumn{2}{c}{VisA} & \multicolumn{2}{c}{MVTec LOCO} & \multicolumn{2}{c}{BrainMRI} & \multicolumn{2}{c}{LiverCT} & \multicolumn{2}{c}{RESC} & \multicolumn{2}{c}{Mean} \\
\cmidrule(lr){2-3}\cmidrule(lr){4-5}\cmidrule(lr){6-7}\cmidrule(lr){8-9}\cmidrule(lr){10-11}\cmidrule(lr){12-13}\cmidrule(lr){14-15}
 & Img. & Pixel & Img. & Pixel & Img. & Pixel & Img. & Pixel & Img. & Pixel & Img. & Pixel & Img. & Pixel \\
\midrule
PatchCore~\cite{roth2022towards} & 84.0 & 89.9 & 74.8 & 93.4 & 62.0 & 69.8 & 73.2 & 96.0 & 44.9 & 95.6 & 56.3 & 78.2 & 65.9 & 87.2 \\
AnomalyGPT~\cite{gu2024anomalygpt} & 94.1 & 95.3 & 87.4 & \underline{96.2} & 60.4 & 70.3 & 73.1 & 96.0 & 60.3 & 95.8 & 82.4 & 94.0 & 76.3 & 91.3 \\
WinCLIP~\cite{jeong2023winclip} & 93.1 & 95.2 & 83.8 & \underline{96.2} & 58.0 & 58.8 & 55.4 & 86.6 & 60.3 & 94.5 & 72.9 & 87.9 & 70.6 & 86.5 \\
ComAD~\cite{liu2023component} & 57.3 & - & 53.9 & - & 62.2 & - & 33.3 & - & 45.0 & - & 73.5 & - & 54.2 & - \\
UniAD~\cite{you2022unified} & 70.3 & 90.7 & 61.3 & 90.3 & 50.9 & 70.6 & 50.0 & 93.6 & 35.0 & 88.5 & 53.5 & 80.7 & 53.5 & 85.7 \\
MedCLIP~\cite{wang2022medclip} & 75.2 & 79.1 & 69.0 & 88.2 & 54.9 & 69.1 & 69.7 & 91.7 & 40.5 & 93.8 & 66.9 & 91.5 & 62.7 & 85.6 \\
UniVAD~\cite{gu2025univad} & 97.8 & \textbf{96.5} & 93.5 & \textbf{98.2} & 71.0 & \textbf{75.1} & 80.2 & \textbf{96.8} & 70.0 & 96.3 & 85.5 & 94.9 & 83.0 & \underline{93.0} \\
\rowcolor{methodgray}
UniVAD v2 (1N-shot) & \underline{98.1} & \underline{96.4} & \underline{93.8} & \textbf{98.2} & \underline{71.5} & \underline{75.0} & \underline{82.4} & \underline{96.7} & \underline{74.5} & \underline{96.8} & \underline{86.9} & \underline{95.0} & \underline{84.5} & \underline{93.0} \\
\rowcolor{methodgray}
UniVAD v2 (1N+1A-shot) & \textbf{98.3} & \textbf{96.5} & \textbf{94.6} & \textbf{98.2} & \textbf{72.1} & 74.9 & \textbf{85.8} & \textbf{96.8} & \textbf{75.6} & \textbf{97.0} & \textbf{87.5} & \textbf{95.2} & \textbf{85.7} & \textbf{93.1} \\
\bottomrule
\end{tabular}
}
\end{table*}

\section{Experiments}
\label{sec:experiments}

\subsection{Experimental Settings}

\subsubsection{Datasets}
We evaluate UniVAD v2 on six datasets spanning industrial, logical, and medical anomaly detection. For industrial inspection, we use MVTec-AD~\cite{bergmann2019mvtec} and VisA~\cite{zou2022spot}, which cover diverse texture and object defects. For logical anomaly detection, we use MVTec LOCO~\cite{bergmann2022beyond}, where anomalies often arise from incorrect compositions of otherwise normal components. For the medical domain, we use BrainMRI~\cite{baid2021rsna}, LiverCT~\cite{bao2024bmad}, and RESC~\cite{hu2019automated}, which represent brain lesion detection, liver CT abnormality localization, and retinal anomaly detection, respectively. All datasets provide pixel-level annotations, allowing us to report both image-level detection and pixel-level localization performance. To evaluate boundary customization, we additionally construct MVTec-AD-SS from the abnormal samples in MVTec-AD. The split separates mild anomalies \(G_1\) from severe anomalies \(G_2\) according to abnormal-region area, anomaly saliency, and human verification. \(G_1\) accounts for roughly 20\% of abnormal images, and \(G_2\) contains the remaining 80\%.

\subsubsection{Evaluation Protocols}
Following UniVAD, we denote a protocol with \(K_n\) normal support images and \(K_a\) abnormal support images as \textit{\(K_n\)N+\(K_a\)A-shot}; when no abnormal support is used, we write \textit{\(K_n\)N-shot}. For fair comparison, we follow the common few-shot evaluation protocol used in RegAD~\cite{huang2022registration}, DictAS~\cite{qu2025dictas}, and related methods. Specifically, for each \(K_n\)N-shot setting, \(K_n\) normal support samples are randomly selected from the target category, and all compared methods are evaluated using the same selected normal supports. Under the \textit{1N-shot} setting, the model is not trained on target-domain data; instead, one normal support image is provided at inference time for each episode. This setting evaluates fair normal-side boundary construction from the same support information used by previous baselines. Under the \textit{1N+1A-shot} setting, one additional abnormal support image is further provided to activate FAR and evaluate the benefit of abnormal-side evidence. For cross-domain transfer, the model trained on MVTec-AD is evaluated on VisA, MVTec LOCO, BrainMRI, LiverCT, and RESC, while the MVTec-AD results are obtained using the model trained on VisA. On MVTec-AD-SS, we directly use VisA-trained weights and do not perform any target-domain retraining. This protocol evaluates whether abnormal references from \(G_1\) or \(G_2\) can induce different inference-time boundaries, including a permissive setting where \(G_1\) mild anomalies are treated as acceptable normal cases and \(G_2\) severe anomalies are treated as abnormal cases. Following established anomaly detection practice, we use image-level AUC and pixel-level AUC to evaluate anomaly detection and anomaly localization, respectively.

\subsubsection{Implementation Details}
In our experiments, the input resolution is set to \(448\times448\), consistent with UniVAD. The model is trained with a batch size of 8, a learning rate of \(1\times10^{-5}\), and the pairwise ranking loss in Eq.~\eqref{eq:rank_loss} for 50 epochs. During inference, the dense anomaly map is upsampled to the image resolution, and the image-level score is obtained from the peak response of the final map. All experiments are conducted on a single NVIDIA RTX A6000 GPU.

\subsection{Comparison with State-of-the-Art}

We compare UniVAD v2 with representative anomaly detection methods from different domains, including PatchCore~\cite{roth2022towards}, AnomalyGPT~\cite{gu2024anomalygpt}, WinCLIP~\cite{jeong2023winclip}, ComAD~\cite{liu2023component}, UniAD~\cite{you2022unified}, MedCLIP~\cite{wang2022medclip}, and UniVAD~\cite{gu2025univad}. Table~\ref{tab:sota_main} reports the results under the fixed normal-support protocol described above. Since all methods receive the same selected normal supports in each target category, the comparison isolates the ability of each method to construct a decision boundary from identical normal-support evidence. Under the 1N-shot setting, UniVAD v2 raises the six-dataset mean image-level AUC from 83.0\% to 84.5\% relative to UniVAD while maintaining the same 93.0\% mean pixel-level AUC. This establishes a fair state-of-the-art comparison against existing normal-support methods and shows that ACRRM and OTRM strengthen normal-side boundary construction even without abnormal references. The gains are most visible on the medical benchmarks, where image-level AUC increases from 80.2\% to 82.4\% on BrainMRI, from 70.0\% to 74.5\% on LiverCT, and from 85.5\% to 86.9\% on RESC. Under the reference-assisted 1N+1A-shot setting, UniVAD v2 further raises the mean image-level and pixel-level AUCs to 85.7\% and 93.1\%, respectively. The largest additional gain appears on BrainMRI, where image-level AUC increases from 82.4\% to 85.8\%. These results separate the two sources of improvement, since normal-side coordination improves cross-domain transfer in the fixed 1N-shot regime, while abnormal supports provide complementary abnormal-side evidence for constructing a more flexible support-conditioned boundary at inference time.

\begin{table}[t]
\centering
\caption{Average single-sample inference time in milliseconds.}
\label{tab:efficiency}
\resizebox{0.85\linewidth}{!}{
\begin{tabular}{lc}
\toprule
Method & Inference Time (ms) \\
\midrule
UniVAD~\cite{gu2025univad} & 658.5 \\
UniVAD v2 (1N+1A-shot) & 679.3 \\
\bottomrule
\end{tabular}
}
\end{table}

Table~\ref{tab:efficiency} further shows that UniVAD v2 preserves favorable inference efficiency. Its average inference time is 679.3 ms per sample, only 20.8 ms slower than UniVAD. This result indicates that the added coordination of retrieval and relational modeling together with abnormal-reference modeling introduces only marginal overhead relative to the accuracy gains reported in Table~\ref{tab:sota_main}.

\subsection{Support-conditioned Boundary Customization}
\begin{figure}[t]
    \centering
    \includegraphics[width=\linewidth]{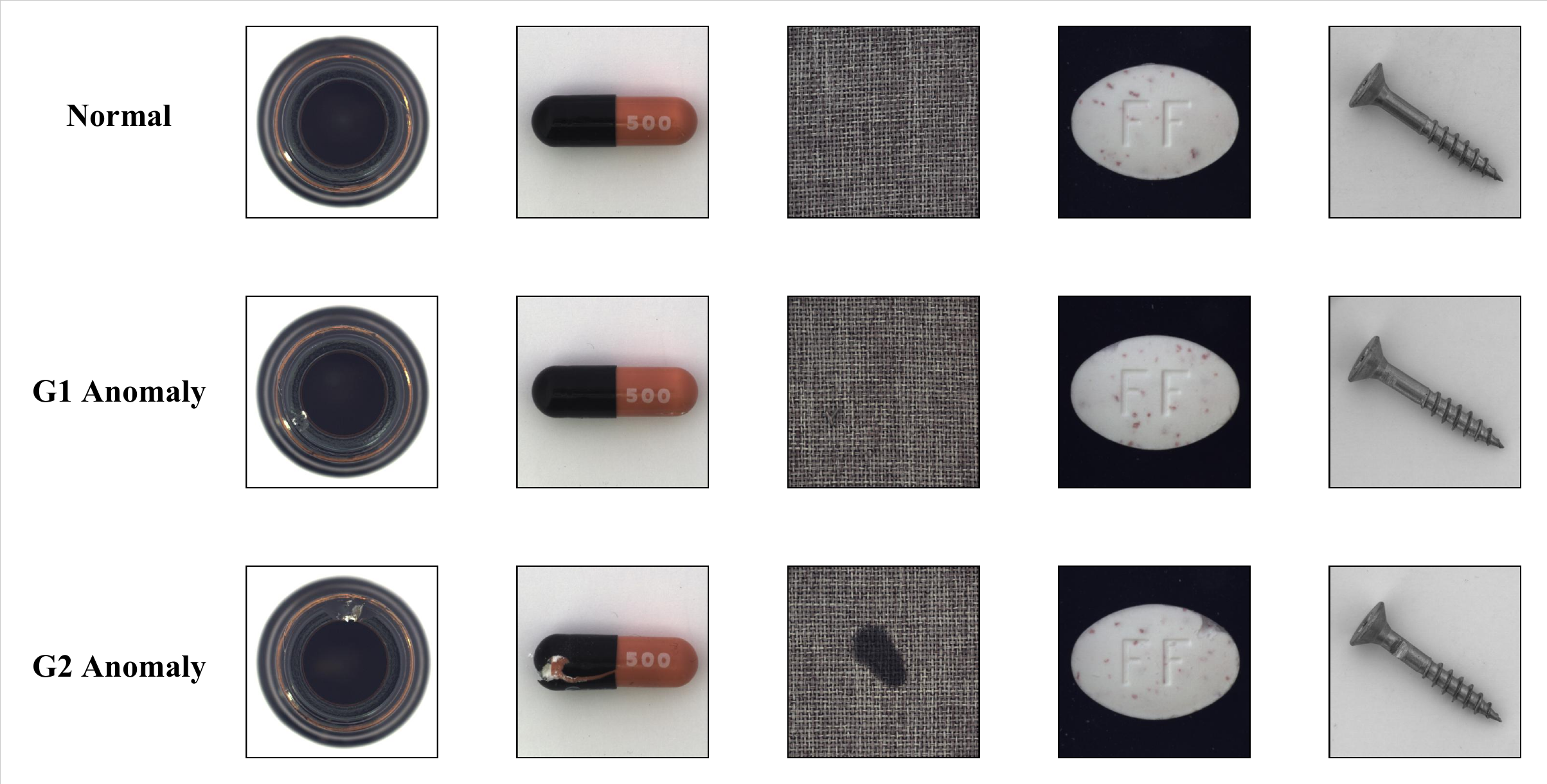}
    \caption{Illustration of MVTec-AD-SS. Each column shows one object category with normal samples, mild \(G_1\) anomalies, and severe \(G_2\) anomalies. \(G_1\) contains subtle deviations that may be acceptable under a permissive tolerance, while \(G_2\) contains stronger defects.}
    \label{fig:mvtec_ss}
\end{figure}

We evaluate whether FAR can adjust the anomaly scoring behavior according to the abnormal references provided at inference time. This experiment is conducted on MVTec-AD-SS and directly uses the model trained on VisA. Different from the fixed-support state-of-the-art comparison in Table~\ref{tab:sota_main}, this analysis changes the abnormal-reference condition to probe how the inferred boundary responds to deployment-specific tolerance requirements. The operational definition of abnormality is also changed, where mild defects in \(G_1\) can be treated as acceptable normal cases when the target application only needs to reject severe defects in \(G_2\). Fig.~\ref{fig:mvtec_ss} provides representative examples of the normal, \(G_1\), and \(G_2\) groups. We use 4 abnormal references in this evaluation to obtain a more stable estimate of the abnormal side, since a single abnormal sample may reflect an instance-specific defect pattern rather than the intended severity level. This setting remains few-shot and uses the abnormal references only at inference time, so it still evaluates boundary customization without target-domain retraining.

\begin{table}[t]
\centering
\caption{Boundary-customization evaluation on MVTec-AD-SS. \(G_1\) mild anomalies are treated as acceptable samples, \(G_2\) severe anomalies are treated as abnormal samples, and UniVAD v2 uses four abnormal references sampled from \(G_2\). No method is retrained on this split. AUC values are reported in \%.}
\label{tab:severity_boundary}
\resizebox{0.95\linewidth}{!}{
\begin{tabular}{lcc}
\toprule
Method & Image-AUC & Pixel-AUC \\
\midrule
PatchCore~\cite{roth2022towards} & 88.5 & 95.8 \\
UniVAD~\cite{gu2025univad} & 91.6 & 95.9 \\
\rowcolor{methodgray}
Ours (1N+4A-shot) & \textbf{96.2} & \textbf{96.9} \\
\bottomrule
\end{tabular}
}
\end{table}



\begin{table}[t]
\centering
\caption{Average FAR scores on MVTec-AD-SS under different abnormal-reference groups. \(G_1\) references correspond to a stricter boundary, while \(G_2\) references correspond to a more permissive boundary that focuses on severe defects.}
\label{tab:severity_scores}
\resizebox{0.95\linewidth}{!}{
\begin{tabular}{lcc}
\toprule
Abnormal references & \(G_1\) mean score & \(G_2\) mean score \\
\midrule
\(G_1\) as references & 0.61~$\green{\pmb{\uparrow}}$& 0.76~$\red{\pmb{\downarrow}}$ \\
\(G_2\) as references & 0.57~$\red{\pmb{\downarrow}}$ & 0.82~$\green{\pmb{\uparrow}}$ \\
\bottomrule
\end{tabular}
}
\end{table}

\begin{table*}[t]
\centering
\caption{Effect of retrieval and relational modeling integration in the 1N-shot setting. Image-level AUC (Img.) and pixel-level AUC (Pixel) are reported for each dataset together with their six-dataset mean (Mean) (\%). The joint row corresponds to UniVAD v2 (1N-shot) excludes FAR.}
\label{tab:ablation_rr}
\resizebox{\textwidth}{!}{
\begin{tabular}{lcccccccccccccc}
\toprule
\multirow{2}{*}{Variant} & \multicolumn{2}{c}{MVTec-AD} & \multicolumn{2}{c}{VisA} & \multicolumn{2}{c}{MVTec LOCO} & \multicolumn{2}{c}{BrainMRI} & \multicolumn{2}{c}{LiverCT} & \multicolumn{2}{c}{RESC} & \multicolumn{2}{c}{Mean} \\
\cmidrule(lr){2-3}\cmidrule(lr){4-5}\cmidrule(lr){6-7}\cmidrule(lr){8-9}\cmidrule(lr){10-11}\cmidrule(lr){12-13}\cmidrule(lr){14-15}
& Img. & Pixel & Img. & Pixel & Img. & Pixel & Img. & Pixel & Img. & Pixel & Img. & Pixel & Img. & Pixel \\
\midrule
Retrieval branch only & 97.8 & \textbf{96.5} & 93.5 & \textbf{98.2} & 71.0 & \textbf{75.0} & 80.2 & \textbf{96.8} & 70.0 & 96.3 & 85.5 & 94.9 & 83.0 & \textbf{93.0} \\
Relational modeling branch only & \textbf{98.1} & 96.4 & \textbf{94.5} & 97.7 & 71.2 & 74.6 & 80.4 & 96.6 & 74.3 & 96.2 & 85.8 & \textbf{95.0} & 84.1 & 92.8 \\
\rowcolor{methodgray}
Joint retrieval + relational modeling (1N-shot) & \textbf{98.1} & 96.4 & 93.8 & \textbf{98.2} & \textbf{71.5} & \textbf{75.0} & \textbf{82.4} & 96.7 & \textbf{74.5} & \textbf{96.8} & \textbf{86.9} & \textbf{95.0} & \textbf{84.5} & \textbf{93.0} \\
\bottomrule
\end{tabular}
}
\end{table*}

\begin{figure*}[t]
    \centering
    \includegraphics[width=\linewidth]{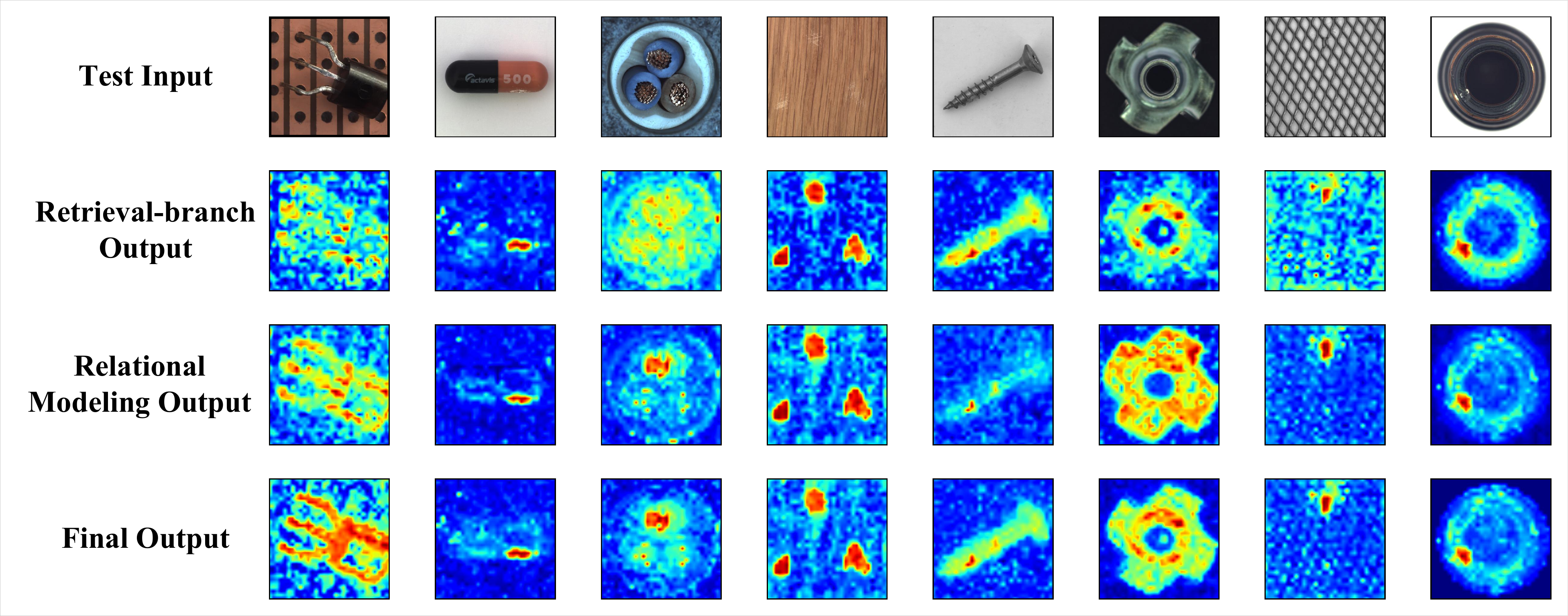}
    \caption{Qualitative visualization of branch complementarity on diverse anomaly types. Each column shows a test input together with the anomaly maps from the retrieval branch, the relational modeling branch, and the final fused output. The retrieval branch tends to produce sharper responses on compact appearance defects, whereas the relational modeling branch provides more complete and spatially coherent activation for structurally extended anomalies. The final prediction combines local sensitivity with global consistency and localizes the anomalous regions more reliably across heterogeneous cases.}
    \label{fig:branch_visualization}
\end{figure*}

Table~\ref{tab:severity_boundary} reports the permissive-boundary setting, where abnormal references are sampled from the severe group \(G_2\) and \(G_1\) mild anomalies are treated as acceptable samples. UniVAD v2 reaches 96.2\% image-level AUC and 96.9\% pixel-level AUC, outperforming PatchCore by 7.7 and 1.1 points and UniVAD by 4.6 and 1.0 points, respectively. This result shows that FAR can use severe abnormal references to construct a boundary aligned with the target industrial tolerance, instead of assigning the same abnormal status to every visible deviation. Table~\ref{tab:severity_scores} further confirms that the scoring behavior is conditioned on the selected abnormal references. When \(G_1\) mild anomalies are used as references, the mean FAR score of \(G_1\) samples increases from 0.57 to 0.61, which is consistent with a stricter effective boundary that rejects subtle defects. When \(G_2\) severe anomalies are used as references, the \(G_1\) score decreases to 0.57 while the \(G_2\) score increases to 0.82, yielding a wider separation between acceptable mild deviations and severe defects. These results support the key claim of FAR that abnormal references provide controllable negative-side evidence that adjusts the inference-time anomaly scoring behavior without retraining on the target category.

\subsection{Ablation Study}
We organize the ablation study around 4 factors, including 1) the complementarity between retrieval and relational modeling for normal-side boundary construction, 2) the contribution of FAR beyond the normal-only setting and the importance of its residual design, 3) the effectiveness of ACRRM relative to simpler fusion strategies, and 4) the insertion scope of OTRM within the hierarchical branches. Together, these studies clarify how each component contributes to the final performance and how normal-side and abnormal-side evidence shape the decision process.

\subsubsection{Retrieval and Relational Modeling Integration}
We first examine whether the relational modeling branch is effective on its own and whether combining it with retrieval is necessary for normal-side boundary construction. Table~\ref{tab:ablation_rr} compares a retrieval-only variant, a relational-modeling-only variant, and their integrated version without FAR.

The results reveal a clear complementarity between the two branches. Compared with the retrieval-only variant, the relational-modeling-only variant raises the six-dataset mean image-level AUC from 83.0\% to 84.1\%, with particularly visible gains on VisA, LiverCT, and RESC, indicating that the OTRM branch better captures global context and category transfer. However, relational modeling alone is slightly less stable for pixel-level localization on several datasets, showing that the retrieval branch still provides stronger local correspondence. When the two branches are coordinated in the 1N-shot setting, the joint normal-side model further reaches 84.5\% mean image-level AUC and 93.0\% mean pixel-level AUC. This result confirms that retrieval and relational modeling provide complementary evidence for normal-side boundary construction before any abnormal reference is introduced. Fig.~\ref{fig:branch_visualization} further shows that branch preference is instance-dependent rather than fixed. On compact and appearance-driven defects, the retrieval branch often produces sharper local peaks and cleaner contrast around the abnormal region. On spatially extended or structurally organized anomalies, the relational modeling branch tends to generate more complete and spatially coherent responses because it models the support-query relation more holistically. After ACRRM fuses the two branches, the final prediction preserves localized high-confidence responses when the anomaly is small, while also covering the full abnormal region when broader structural context is required. This visual evidence is consistent with the quantitative results in Table~\ref{tab:ablation_rr}, where neither branch is uniformly dominant across all cases, and their adaptive coordination yields the strongest overall performance.

\subsubsection{Few-Shot Abnormal Reference Module}
We next evaluate whether abnormal supports provide useful boundary-side information beyond normal supports and which parts of FAR are most important. Table~\ref{tab:ablation_far} compares the normal-only 1N-shot setting, several degraded FAR variants, and the full 1N+1A-shot formulation.

\begin{table*}[t]
\centering
\caption{Effect of the Few-Shot Abnormal Reference module (FAR) and its design choices. Image-level AUC (Img.) and pixel-level AUC (Pixel) are reported for each dataset together with their six-dataset mean (Mean) (\%).}
\label{tab:ablation_far}
\resizebox{\textwidth}{!}{
\begin{tabular}{lcccccccccccccc}
\toprule
\multirow{2}{*}{Variant} & \multicolumn{2}{c}{MVTec-AD} & \multicolumn{2}{c}{VisA} & \multicolumn{2}{c}{MVTec LOCO} & \multicolumn{2}{c}{BrainMRI} & \multicolumn{2}{c}{LiverCT} & \multicolumn{2}{c}{RESC} & \multicolumn{2}{c}{Mean} \\
\cmidrule(lr){2-3}\cmidrule(lr){4-5}\cmidrule(lr){6-7}\cmidrule(lr){8-9}\cmidrule(lr){10-11}\cmidrule(lr){12-13}\cmidrule(lr){14-15}
& Img. & Pixel & Img. & Pixel & Img. & Pixel & Img. & Pixel & Img. & Pixel & Img. & Pixel & Img. & Pixel \\
\midrule
\rowcolor{methodgray}
1N-shot & 98.1 & 96.4 & 93.8 & \textbf{98.2} & 71.5 & \textbf{75.0} & 82.4 & 96.7 & 74.5 & 96.8 & 86.9 & 95.0 & 84.5 & 93.0 \\
\,+ abnormal-memory & 97.8 & 96.4 & 93.9 & 97.9 & 69.4 & 74.8 & 84.5 & 96.4 & 75.6 & 96.8 & 86.5 & 95.0 & 84.6 & 92.9 \\
\,+ abnormal w/o residual & 97.7 & 96.2 & 94.1 & 97.9 & 69.8 & 74.8 & 82.3 & 96.5 & 71.9 & \textbf{97.0} & 86.4 & 94.8 & 83.7 & 92.9 \\
\,+ query w/o residual & 98.0 & 96.4 & 93.6 & 97.7 & 70.1 & 74.5 & 80.3 & 96.4 & 74.9 & 96.7 & 84.2 & 94.5 & 83.5 & 92.7 \\
\rowcolor{methodgray}
+ full FAR & \textbf{98.3} & \textbf{96.5} & \textbf{94.6} & \textbf{98.2} & \textbf{72.1} & 74.9 & \textbf{85.8} & \textbf{96.8} & \textbf{75.6} & \textbf{97.0} & \textbf{87.5} & \textbf{95.2} & \textbf{85.7} & \textbf{93.1} \\
\bottomrule
\end{tabular}
}
\end{table*}

Adding one abnormal support image with the full FAR design improves the six-dataset mean image-level AUC from 84.5\% to 85.7\% and yields the highest or tied-highest value on 13 metrics among the 14 reported image-level and pixel-level AUC values over the six datasets and their mean. This isolates the main benefit of abnormal-side boundary evidence beyond the normal-side model analyzed in Table~\ref{tab:ablation_rr}. Replacing FAR with a direct abnormal-memory branch still helps on some datasets, but its mean image-level AUC remains 1.1 points below the full design, showing that raw abnormal-feature storage does not provide a reliable boundary constraint. Removing residual mining on the abnormal side or the query side causes a larger drop, reducing the mean image-level AUC to 83.7\% and 83.5\%, respectively. These results confirm that the residual formulation is the key ingredient in FAR. It suppresses normal overlap in both abnormal supports and query features, converts abnormal references into category-agnostic deviations from the current normal manifold, and provides more reliable abnormal-side evidence for unseen categories.

\subsubsection{Effect of ACRRM}
We then assess whether the gain comes from the proposed coordination mechanism or from replacing the fixed fusion rule with simpler fusion strategies. Table~\ref{tab:ablation_arrcm} compares ACRRM with direct summation and a simple MLP gate.

\begin{table*}[t]
\centering
\caption{Effect of different fusion mechanisms. Image-level AUC (Img.) and pixel-level AUC (Pixel) are reported for each dataset together with their six-dataset mean (Mean) (\%).}
\label{tab:ablation_arrcm}
\resizebox{\textwidth}{!}{
\begin{tabular}{lcccccccccccccc}
\toprule
\multirow{2}{*}{Fusion mechanism} & \multicolumn{2}{c}{MVTec-AD} & \multicolumn{2}{c}{VisA} & \multicolumn{2}{c}{MVTec LOCO} & \multicolumn{2}{c}{BrainMRI} & \multicolumn{2}{c}{LiverCT} & \multicolumn{2}{c}{RESC} & \multicolumn{2}{c}{Mean} \\
\cmidrule(lr){2-3}\cmidrule(lr){4-5}\cmidrule(lr){6-7}\cmidrule(lr){8-9}\cmidrule(lr){10-11}\cmidrule(lr){12-13}\cmidrule(lr){14-15}
& Img. & Pixel & Img. & Pixel & Img. & Pixel & Img. & Pixel & Img. & Pixel & Img. & Pixel & Img. & Pixel \\
\midrule
Direct summation & 97.7 & \textbf{96.5} & 93.6 & 97.7 & 71.0 & 74.2 & 83.4 & 96.5 & 71.7 & 96.2 & 85.5 & 95.1 & 83.8 & 92.7 \\
MLP-based fusion & 98.2 & 96.4 & 94.0 & 97.3 & 71.3 & 73.6 & 81.5 & 96.6 & 75.3 & 96.3 & 85.9 & 94.9 & 84.4 & 92.5 \\
\rowcolor{methodgray}
ACRRM-based fusion & \textbf{98.3} & \textbf{96.5} & \textbf{94.6} & \textbf{98.2} & \textbf{72.1} & \textbf{74.9} & \textbf{85.8} & \textbf{96.8} & \textbf{75.6} & \textbf{97.0} & \textbf{87.5} & \textbf{95.2} & \textbf{85.7} & \textbf{93.1} \\
\bottomrule
\end{tabular}
}
\end{table*}

The ACRRM-based fusion model consistently outperforms both direct summation and the MLP-based fusion baseline. Compared with direct summation, ACRRM improves the mean image-level AUC from 83.8\% to 85.7\% and the mean pixel-level AUC from 92.7\% to 93.1\%. Compared with the learnable MLP gate, ACRRM further raises the mean image-level AUC from 84.4\% to 85.7\% and the mean pixel-level AUC from 92.5\% to 93.1\%, indicating that the gain is not attributable to a generic fusion layer alone. Instead, the episode-conditioned coordination mechanism in ACRRM provides a more reliable way to decide when retrieval evidence should dominate and when relational-modeling evidence should be emphasized.

\subsubsection{Effect of OTRM Insertion Scope}
Finally, we study how OTRM should be inserted into the framework. Table~\ref{tab:ablation_otrm} compares progressively broader OTRM insertion scopes, ranging from only the local branch to the full local-part-component configuration.

\begin{table*}[t]
\centering
\caption{Effect of different OTRM insertion scopes. Image-level AUC (Img.) and pixel-level AUC (Pixel) are reported for each dataset together with their six-dataset mean (Mean) (\%). ``Local-Only,'' ``Local + Part,'' and ``Local + Part + Component'' indicate that OTRM is inserted into the corresponding branches on top of the retrieval backbone.}
\label{tab:ablation_otrm}
\resizebox{\textwidth}{!}{
\begin{tabular}{lcccccccccccccc}
\toprule
\multirow{2}{*}{OTRM variant} & \multicolumn{2}{c}{MVTec-AD} & \multicolumn{2}{c}{VisA} & \multicolumn{2}{c}{MVTec LOCO} & \multicolumn{2}{c}{BrainMRI} & \multicolumn{2}{c}{LiverCT} & \multicolumn{2}{c}{RESC} & \multicolumn{2}{c}{Mean} \\
\cmidrule(lr){2-3}\cmidrule(lr){4-5}\cmidrule(lr){6-7}\cmidrule(lr){8-9}\cmidrule(lr){10-11}\cmidrule(lr){12-13}\cmidrule(lr){14-15}
& Img. & Pixel & Img. & Pixel & Img. & Pixel & Img. & Pixel & Img. & Pixel & Img. & Pixel & Img. & Pixel \\
\midrule
Local-Only & 98.0 & \textbf{96.5} & 94.2 & 98.1 & 71.3 & 74.9 & 83.5 & 96.5 & 73.6 & 96.9 & 86.4 & 95.3 & 84.5 & 93.0 \\
Local + Part & 98.1 & \textbf{96.5} & 94.5 & \textbf{98.2} & 71.5 & \textbf{75.0} & 85.2 & 96.7 & 73.6 & \textbf{97.1} & 87.2 & \textbf{95.4} & 85.0 & \textbf{93.2} \\
\rowcolor{methodgray}
Local + Part + Component (Full OTRM) & \textbf{98.3} & \textbf{96.5} & \textbf{94.6} & \textbf{98.2} & \textbf{72.1} & 74.9 & \textbf{85.8} & \textbf{96.8} & \textbf{75.6} & 97.0 & \textbf{87.5} & 95.2 & \textbf{85.7} & 93.1 \\
\bottomrule
\end{tabular}
}
\end{table*}

Table~\ref{tab:ablation_otrm} shows a clear performance trend as OTRM is inserted into progressively broader branches. With OTRM inserted only into the local branch, the model achieves 84.5\% mean image-level AUC and 93.0\% mean pixel-level AUC. Extending OTRM to both the local and part branches further raises the mean image-level and pixel-level AUCs to 85.0\% and 93.2\%, respectively. The full Local + Part + Component configuration then reaches the strongest image-level performance, achieving an 85.7\% mean image-level AUC and the best image-level result on all six datasets. Notably, the Local + Part variant yields the highest mean pixel-level AUC, whereas Full OTRM yields the highest mean image-level AUC. The reason is that the component branch propagates a consistent anomaly score over the spatial region associated with the detected abnormal component. This behavior is beneficial for image-level anomaly detection because it strengthens the global abnormal evidence of the whole component. However, when the true defect covers only a subset of the component area, assigning elevated scores to the entire component can blur the spatial boundary of the anomaly and slightly reduce pixel-level AUC. These results indicate that transport-based relational modeling is beneficial at multiple semantic levels, and that component-level insertion is particularly important for image-level anomaly discrimination, whereas the Local + Part setting already provides the strongest average localization performance.

\subsection{Additional Analysis}

\subsubsection{Effect of Support Composition and Scaling}

\begin{figure}[t]
    \centering
    \includegraphics[width=\linewidth]{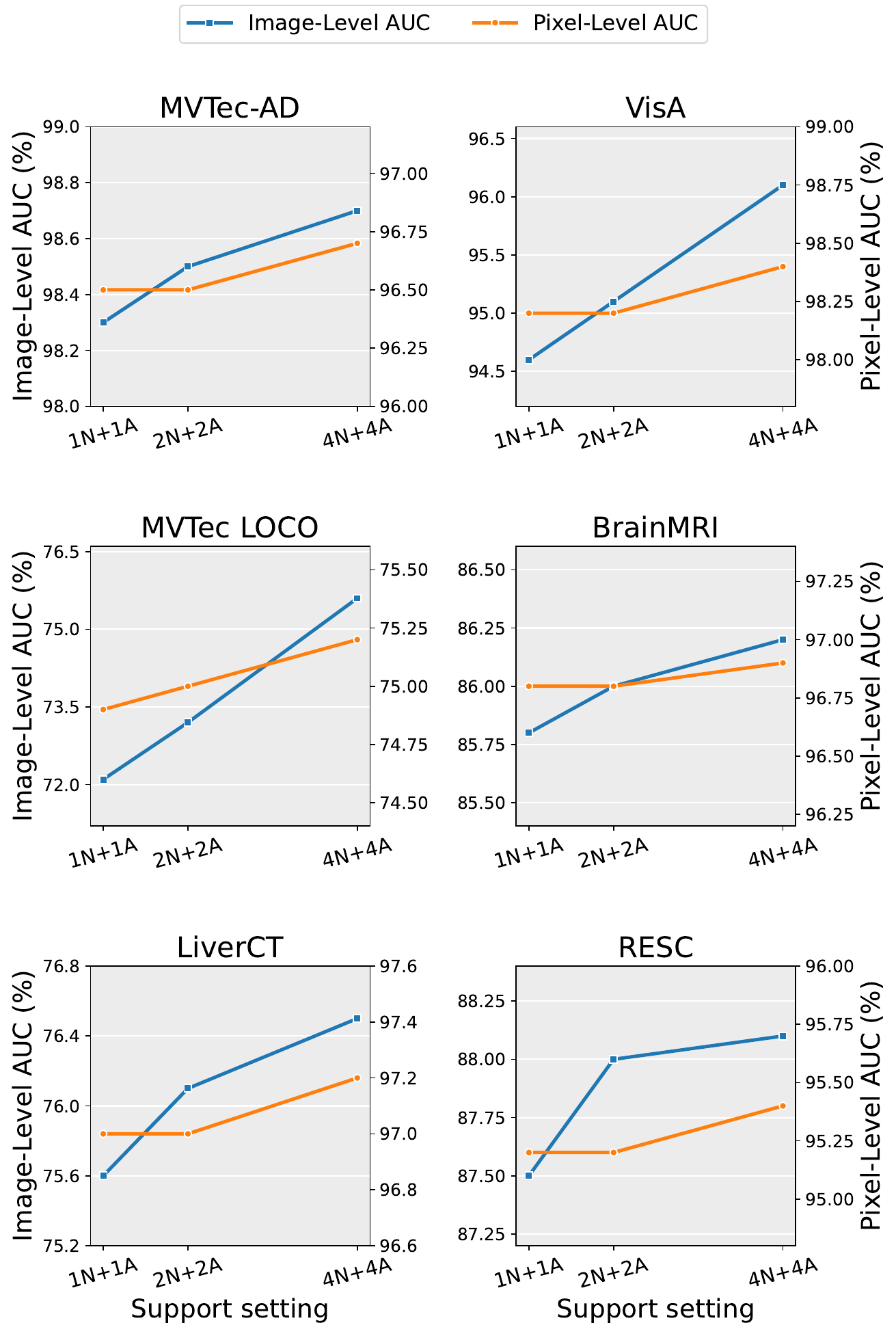}
    \caption{Effect of jointly scaling normal and abnormal support sets. Each subplot reports image-level and pixel-level AUC when the paired support set is expanded from 1N+1A to 2N+2A and 4N+4A.}
    \label{fig:shot_scaling}
\end{figure}

To further evaluate how support-conditioned boundary construction benefits from additional references, we jointly increase the normal and abnormal support sets from 1N+1A to 2N+2A and 4N+4A. Fig.~\ref{fig:shot_scaling} summarizes the resulting image-level and pixel-level AUCs on all six datasets. The six-dataset mean image-level AUC rises steadily from 85.7\% at 1N+1A to 86.2\% at 2N+2A and 86.9\% at 4N+4A, while the mean pixel-level AUC remains stable at 93.1\% and further improves to 93.3\% at 4N+4A. The most visible image-level gains appear on MVTec LOCO and VisA, indicating that additional paired supports strengthen both global relational modeling and support-conditioned boundary evidence. These results show that UniVAD v2 can continue to refine the boundary with larger paired support sets rather than saturating after a single support pair.

\begin{figure}[t]
    \centering
    \includegraphics[width=\linewidth]{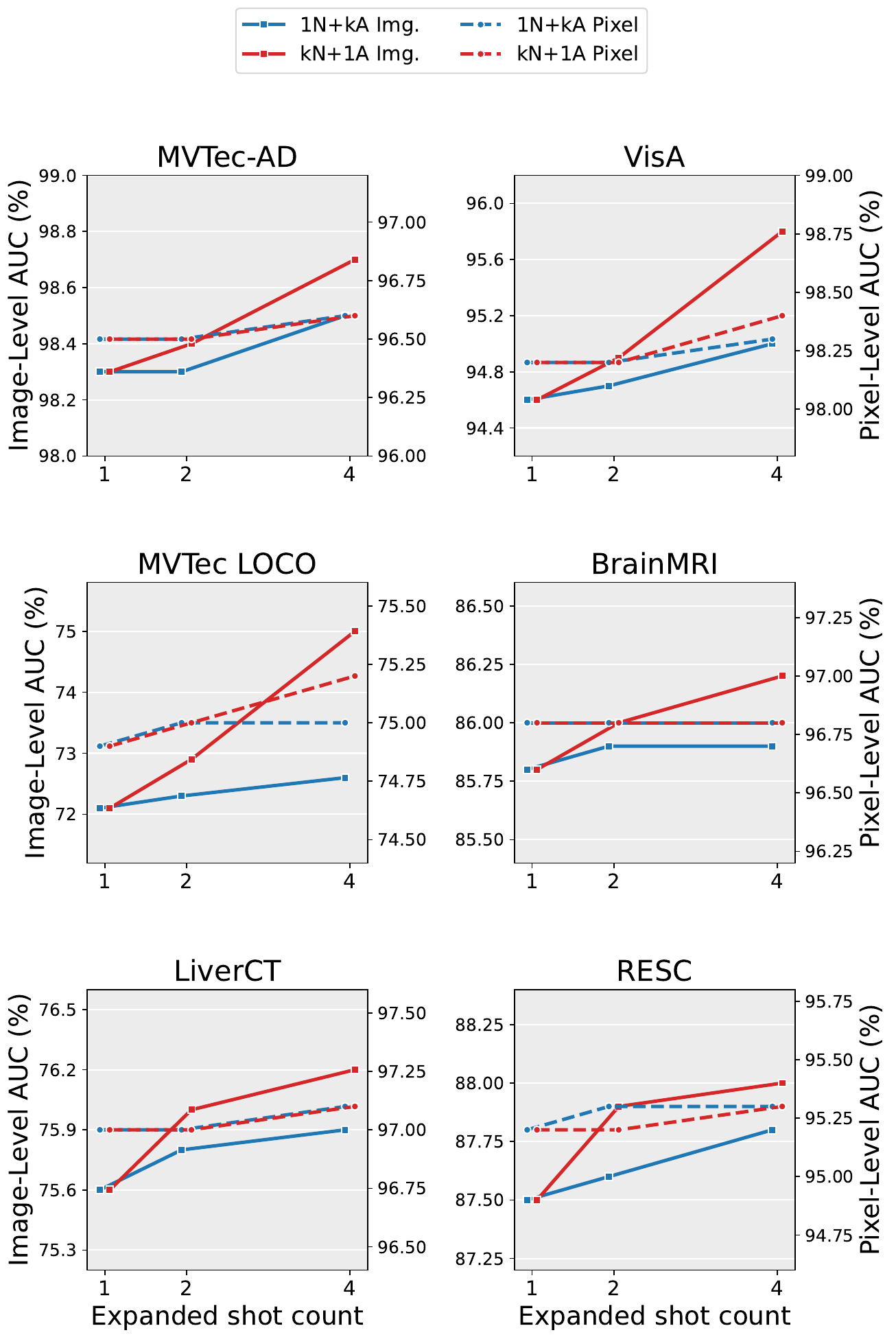}
     \caption{Effect of asymmetric support composition. Each subplot compares two support-expansion strategies by increasing abnormal shots with a fixed 1N support set (1N+kA) or increasing normal shots with a fixed 1A abnormal support set (kN+1A).}
    \label{fig:support_composition}
\end{figure}

We further decouple the two support types by fixing one side and increasing the other. Fig.~\ref{fig:support_composition} compares two asymmetric expansion strategies by increasing abnormal supports with a fixed 1N support set or increasing normal supports with a fixed 1A abnormal support set. UniVAD v2 benefits from both strategies. When the abnormal support count is fixed at 1, the six-dataset mean image-level AUC rises from 85.7\% at 1N+1A to 86.0\% at 2N+1A and 86.7\% at 4N+1A. In contrast, when the normal support count is fixed at 1, increasing abnormal supports yields a more moderate but still consistent improvement from 85.7\% to 85.8\% and 86.0\%. The mean pixel-level AUC remains stable in both settings and reaches 93.2\% or above at larger support counts. These results suggest that normal supports remain the dominant source for defining the acceptable side of the boundary, whereas additional abnormal supports provide abnormal-side evidence for shaping the rejection side.

\subsubsection{Image-Level Score Separation}
\begin{figure}[t]
    \centering
    \includegraphics[width=\linewidth]{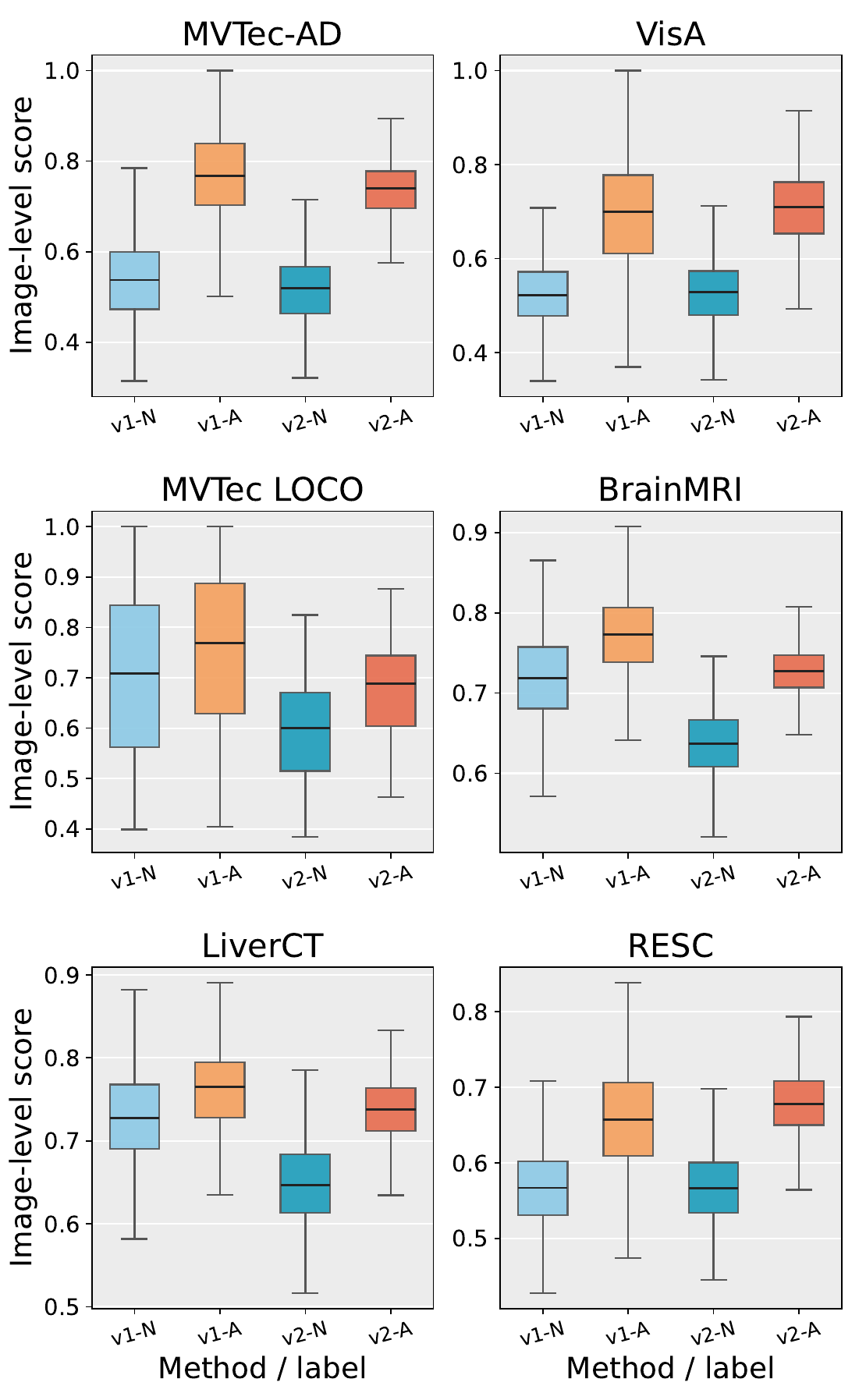}
    \caption{Image-level score separation between normal and abnormal samples for UniVAD and UniVAD v2 under the 1N+1A-shot setting. Each subplot shows the score distributions of normal and abnormal images on one dataset. In the legend, v1 and v2 denote UniVAD and UniVAD v2, respectively, while N and A denote normal and abnormal samples.}
    \label{fig:score_separation}
\end{figure}

To better understand why UniVAD v2 improves image-level discrimination, we compare the image-level score distributions of UniVAD and UniVAD v2 under the 1N+1A-shot setting. Fig.~\ref{fig:score_separation} shows the score distributions.

Fig.~\ref{fig:score_separation} shows that UniVAD v2 significantly enlarges the normal-abnormal score gap on five of the six datasets while maintaining a comparably large separation on MVTec-AD, where the baseline already exhibits strong separability. Averaged over all six datasets, the mean score gap increases from 0.103 for UniVAD to 0.129 for UniVAD v2. The improvement is especially clear on BrainMRI, LiverCT, and MVTec LOCO, where UniVAD v2 suppresses normal scores and raises abnormal scores more consistently. This observation supports the design motivation of UniVAD v2, since normal-side coordination together with abnormal-side reference modeling produces a cleaner support-conditioned decision boundary rather than relying on a fixed boundary learned only from normal training samples.

\subsubsection{ACRRM Weight Allocation}
\begin{figure}[t]
    \centering
    \includegraphics[width=\linewidth]{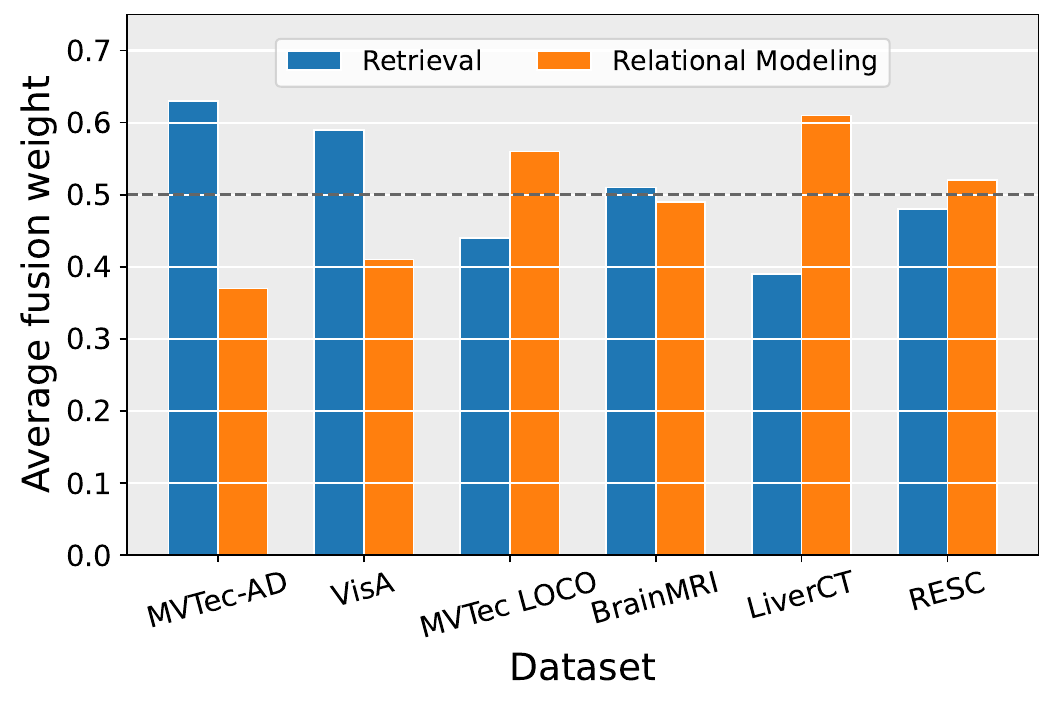}
    \caption{Average branch weights assigned by ACRRM on different datasets. Higher retrieval weights indicate stronger reliance on local support-based matching, while higher relational-modeling weights indicate stronger reliance on global support-query relational modeling.}
    \label{fig:arrcm_allocation}
\end{figure}

To better understand the behavior of the coordination mechanism, we visualize the average fusion weights assigned by ACRRM to the retrieval and relational modeling branches on each dataset in Fig.~\ref{fig:arrcm_allocation}. Fig.~\ref{fig:arrcm_allocation} reveals a dataset-dependent allocation pattern. On MVTec-AD and VisA, ACRRM assigns larger weights to the retrieval branch, which is consistent with the fact that these industrial datasets are dominated by local appearance anomalies. In contrast, on MVTec LOCO, LiverCT, and RESC, the relational modeling branch receives higher weights, suggesting that more global contextual or structural modeling is needed for robust discrimination. BrainMRI exhibits a nearly balanced allocation, implying that both local evidence and global relational modeling are important in that setting. Overall, these results support the design motivation of ACRRM, since UniVAD v2 adapts the relative importance of retrieval and relational modeling according to the characteristics of the target domain instead of relying on a fixed fusion rule.

\subsubsection{Effect of Input Resolution}
\begin{figure}[t]
    \centering
    \includegraphics[width=0.9\linewidth]{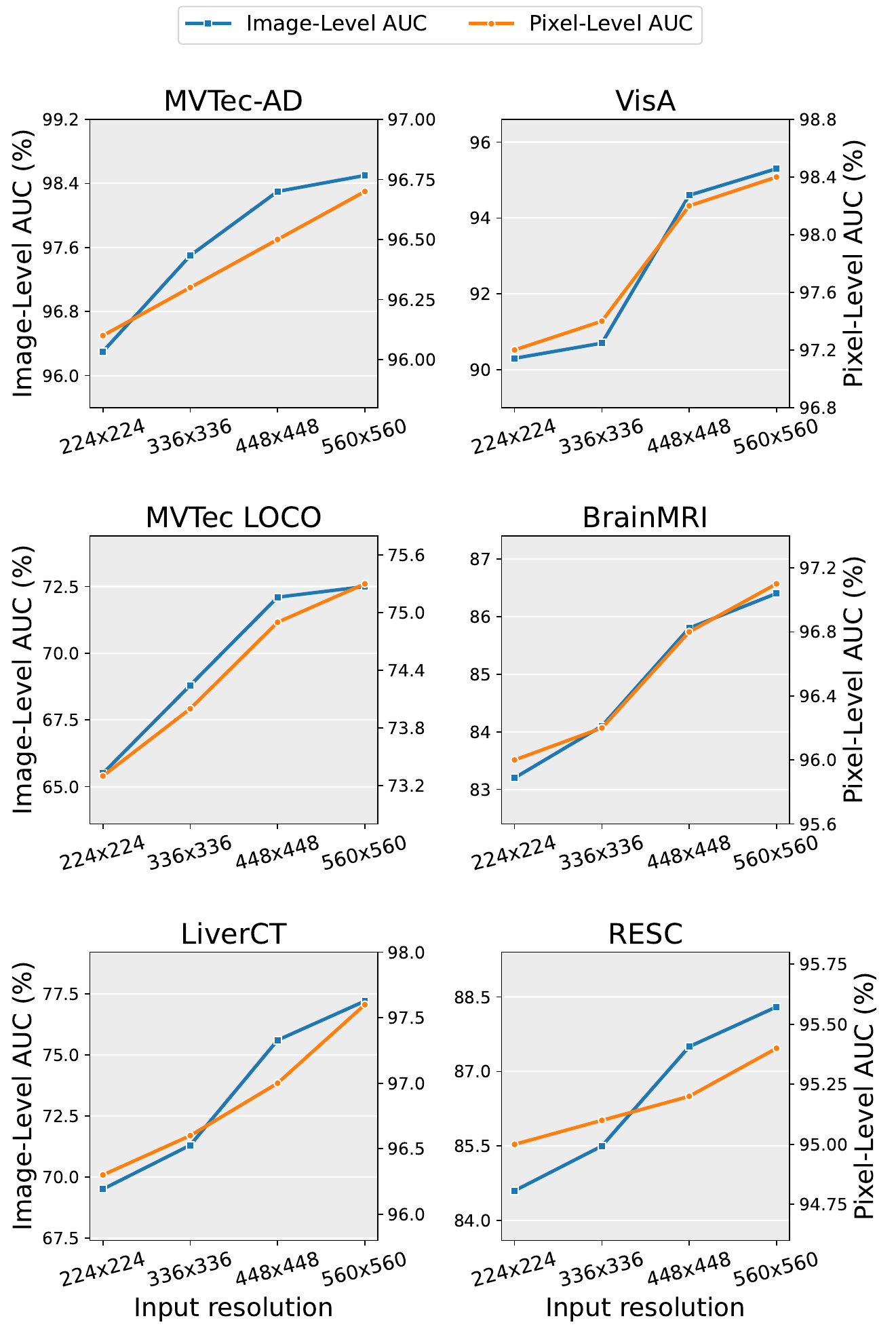}
    \caption{Effect of input resolution. Each subplot reports image-level and pixel-level AUCs as the input resolution is increased from \(224\times224\) to \(560\times560\).}
    \label{fig:resolution_analysis}
\end{figure}

We further study the sensitivity of UniVAD v2 to input resolution. Fig.~\ref{fig:resolution_analysis} reports the image-level and pixel-level AUCs when the input size is varied from \(224\times224\) to \(560\times560\) at inference time.

Fig.~\ref{fig:resolution_analysis} shows that higher input resolution consistently improves the performance of UniVAD v2. The six-dataset mean image-level AUC rises from 81.6\% at \(224\times224\) to 83.0\% at \(336\times336\), 85.7\% at \(448\times448\), and 86.4\% at \(560\times560\), while the mean pixel-level AUC increases from 92.3\% to 92.6\%, 93.1\%, and 93.4\%, respectively. A larger resolution preserves small local defects more clearly and also provides denser spatial evidence for component-level and relational matching. This is useful for support-conditioned boundary construction because both normal-side correspondence and abnormal-side residual evidence depend on the quality of local visual evidence. These results suggest that UniVAD v2 can effectively exploit additional visual detail when the input resolution is increased. In the main experiments, we retain \(448\times448\) as the default setting for fair comparison with UniVAD and to balance performance and computation. The \(560\times560\) results indicate that further gains remain available under a larger inference budget, especially in scenarios where fine spatial defects or structural boundaries are important for reliable anomaly localization.

\section{Limitations and Future Work}
Although UniVAD v2 constructs the inference-time boundary from the current support episode, the quality and representativeness of support samples can still affect the resulting boundary evidence in practical deployment. In the main benchmark comparison, this factor is controlled by fixing the selected normal supports across all compared methods, so the reported state-of-the-art results reflect boundary construction under identical normal-support evidence rather than method-specific support selection. Exhaustively evaluating all possible support combinations is nevertheless infeasible, especially when the number of candidate normal samples varies across datasets and categories. Therefore, following the standard practice in few-shot anomaly detection methods such as RegAD~\cite{huang2022registration} and DictAS~\cite{qu2025dictas}, we randomly select \(K_n\) normal support samples for each \(K_n\)N-shot protocol and evaluate all compared methods on the same selected supports. This protocol ensures a fair comparison under identical support information, while support selection and support aggregation remain important factors for real-world deployment. In addition, MVTec-AD-SS evaluates one practical severity-based tolerance setting, while real deployments may involve more diverse and continuously varying definitions of acceptable defects. Future work will investigate more robust support selection, support aggregation, and broader forms of human- or policy-guided boundary specification.

\section{Conclusion}
We present UniVAD v2, a support-conditioned boundary construction framework for unified visual anomaly detection across industrial, logical, and medical scenarios. Rather than treating unified anomaly detection as a choice between memory-based retrieval and learned normality modeling, UniVAD v2 estimates episode-specific boundary from complementary support information. ACRRM and OTRM strengthen the normal side of this evidence by coordinating local retrieval stability with optimal transport-based global relational modeling, while FAR introduces abnormal-side residual evidence from optional abnormal references. Extensive experiments on six datasets demonstrate that UniVAD v2 improves cross-domain generalization in the fair 1N-shot setting and that retrieval and relational modeling provide complementary normal-side evidence. The 1N+1A-shot results and MVTec-AD-SS analysis further confirm that FAR converts abnormal references into negative-side evidence, enabling stricter or more permissive effective inference-time boundaries from mild or severe abnormal references without retraining. These findings suggest that support-conditioned boundary construction is a practical and extensible formulation for scalable, interpretable, and deployable unified anomaly detection systems.

{
    \bibliographystyle{IEEEtran}
    \bibliography{ref}
}

\begin{IEEEbiography}
[{\includegraphics[width=1in,height=1.25in,clip,keepaspectratio]{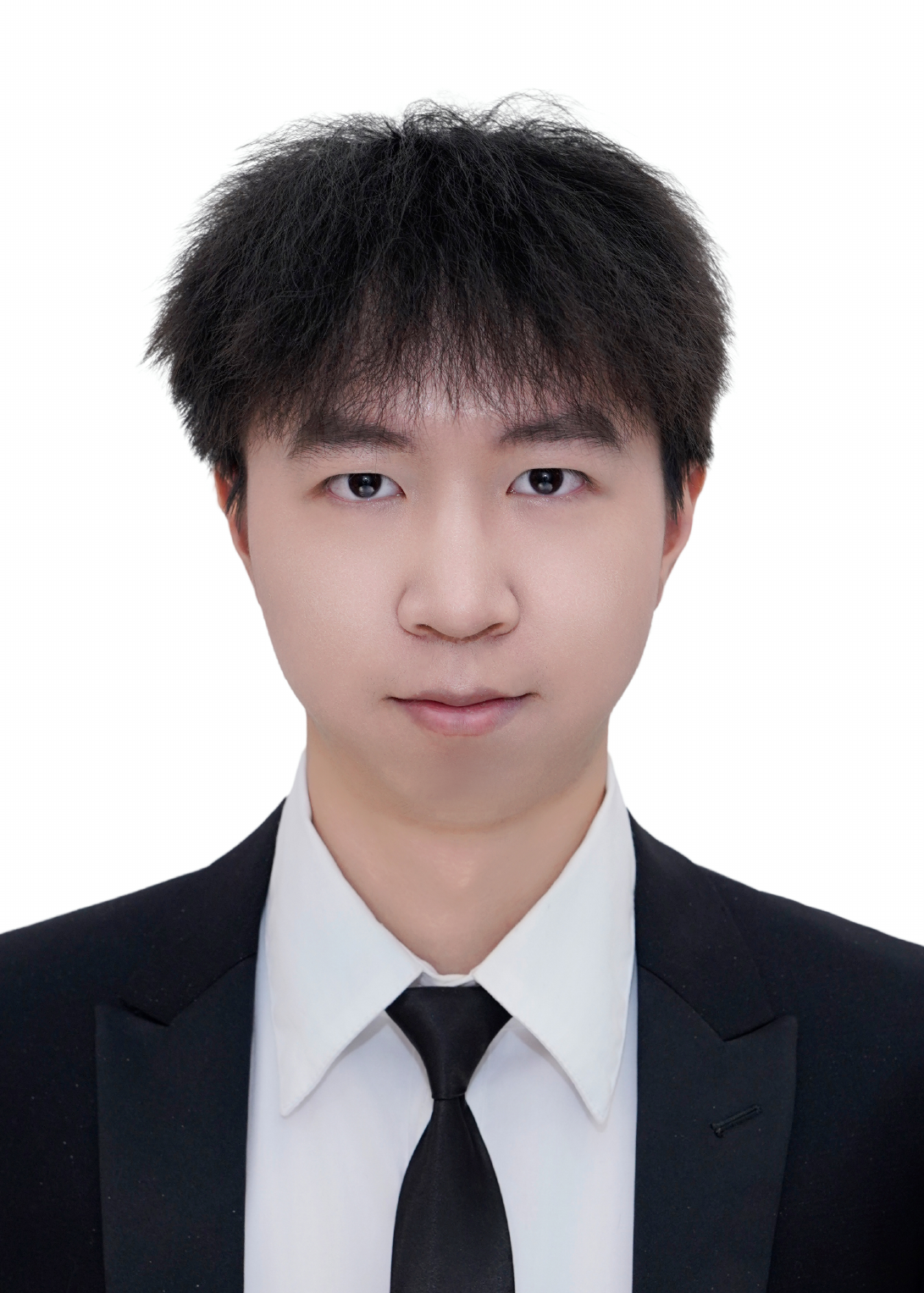}}]{Zhaopeng Gu}
received the B.E. degree from Beijing University of Post and Telecommunication, Beijing, China, in 2023. He is currently pursuing a doctor's degree with the Foundation Model Research Center of the Institute of Automation, Chinese Academy of Sciences, Beijing.

His research interests include computer vision, anomaly detection, and multimodal learning.
\end{IEEEbiography}


\begin{IEEEbiography}
[{\includegraphics[width=1in,height=1.25in,clip,keepaspectratio]{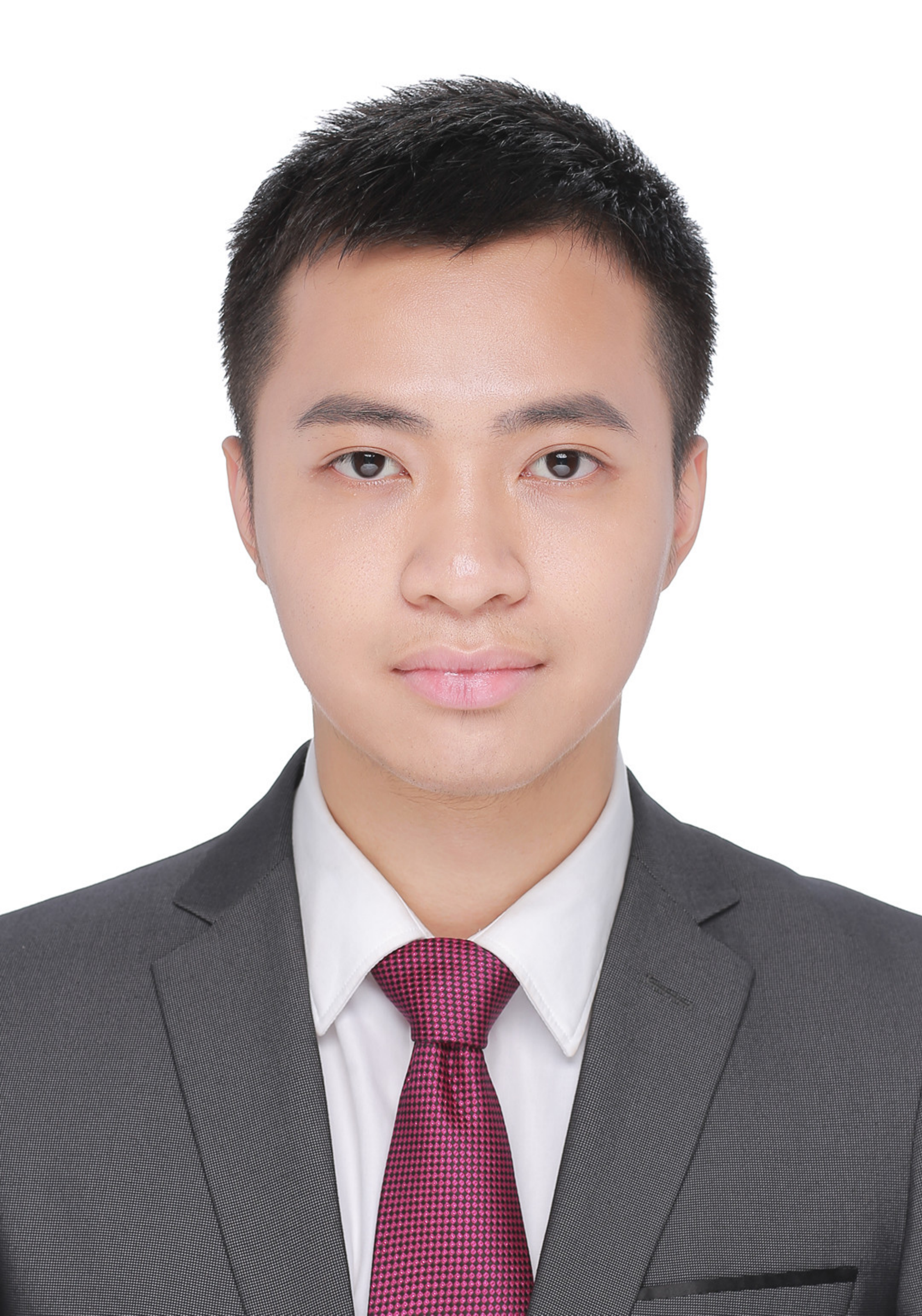}}]{Bingke Zhu}
received the B.E. degree from Beijing
University of Chemical Technology, Beijing, China,
in 2016, and the Ph.D. degree from the University
of Chinese Academy of Sciences, Beijing, in 2021.

He is currently an Assistant Professor with
the Foundation Model Research Center, Institute
of Automation, Chinese Academy of Sciences,
Beijing. His current research interests include pattern recognition and machine learning, image and
video processing, semantic segmentation, and visual
anomaly detection.

\end{IEEEbiography}

\begin{IEEEbiography}
[{\includegraphics[width=1in,height=1.25in,clip,keepaspectratio]{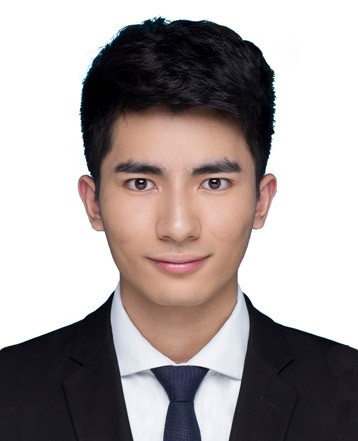}}]{Zhaowen Li}
received the Ph.D. degree from the Foundation Model Research Center, Institute of Automation, Chinese Academy of Sciences, Beijing, in 2024. He is currently a Principal Researcher of Autonomous Driving Research and Development at Huawei. His current research interests include physical artificial intelligence, self-/un-supervised learning, and anomaly detection.

\end{IEEEbiography}

\begin{IEEEbiography}
[{\includegraphics[width=1in,height=1.25in,clip,keepaspectratio]{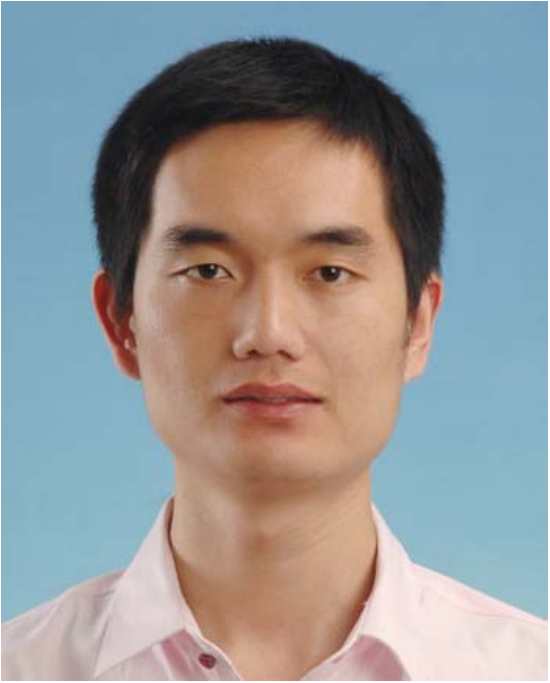}}]{Guibo Zhu}
received the B.E. degree from Wuhan
University, Wuhan, China, in 2009, the M.Sc. degree
from the University of Chinese Academy of Sciences, Beijing, China, in 2013, and the Ph.D. degree
from the National Laboratory of Pattern Recognition, Institute of Automation, Chinese Academy of
Sciences, Beijing, in 2016.

He is currently an Associate Professor with the
Institute of Automation, Chinese Academy of Sciences, Beijing. His current research interests include
computer vision, multimodal pretraining models,
object tracking, and action recognition.
\end{IEEEbiography}

\begin{IEEEbiography}
[{\includegraphics[width=1in,height=1.25in,clip,keepaspectratio]{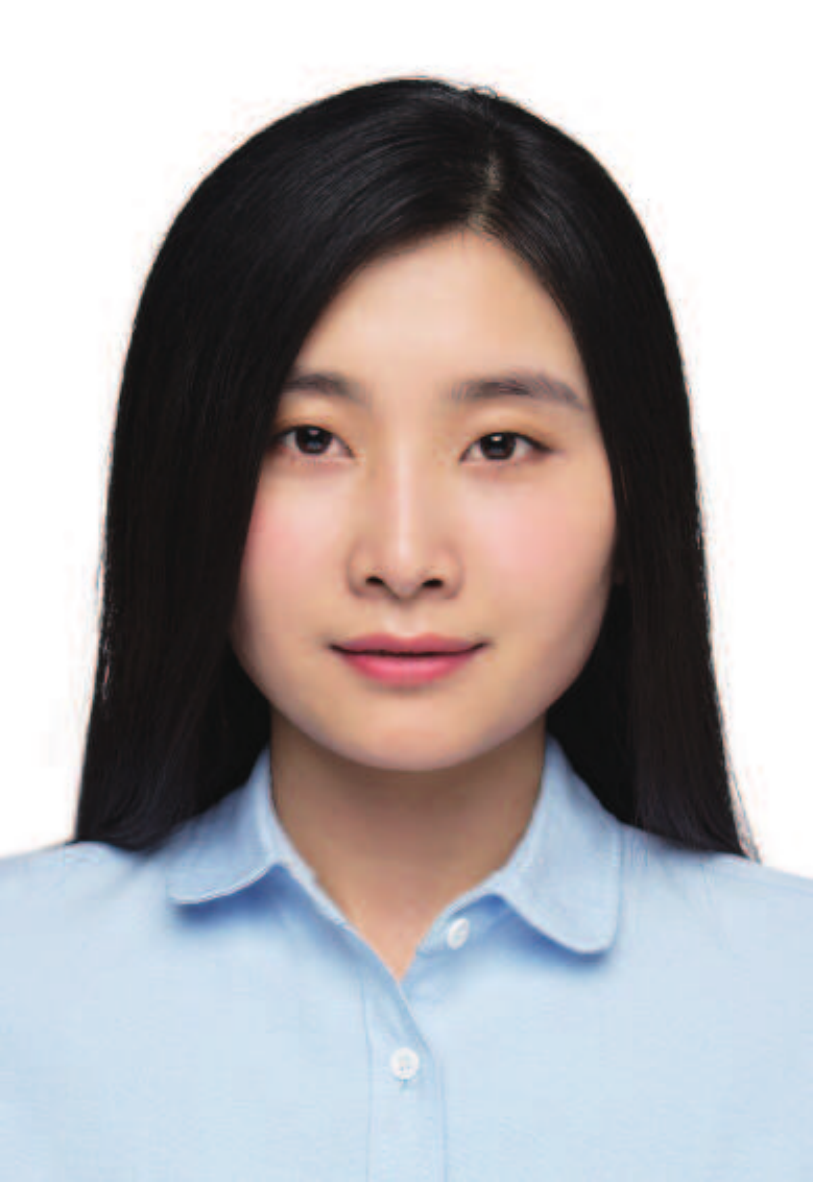}}]{Yingying Chen}
received the B.S. degree from the
Communication University of China, Beijing, China,
in 2013, and the Ph.D. degree from the University
of Chinese Academy of Sciences, Beijing, in 2018.

She is currently an Associate Professor with the
Foundation Model Research Center, Institute of
Automation, Chinese Academy of Sciences, Beijing.
Her current research interests include pattern recognition and machine learning, image and video
processing, and intelligent video surveillance.
\end{IEEEbiography}

\begin{IEEEbiography}
[{\includegraphics[width=1in,height=1.25in,clip,keepaspectratio]{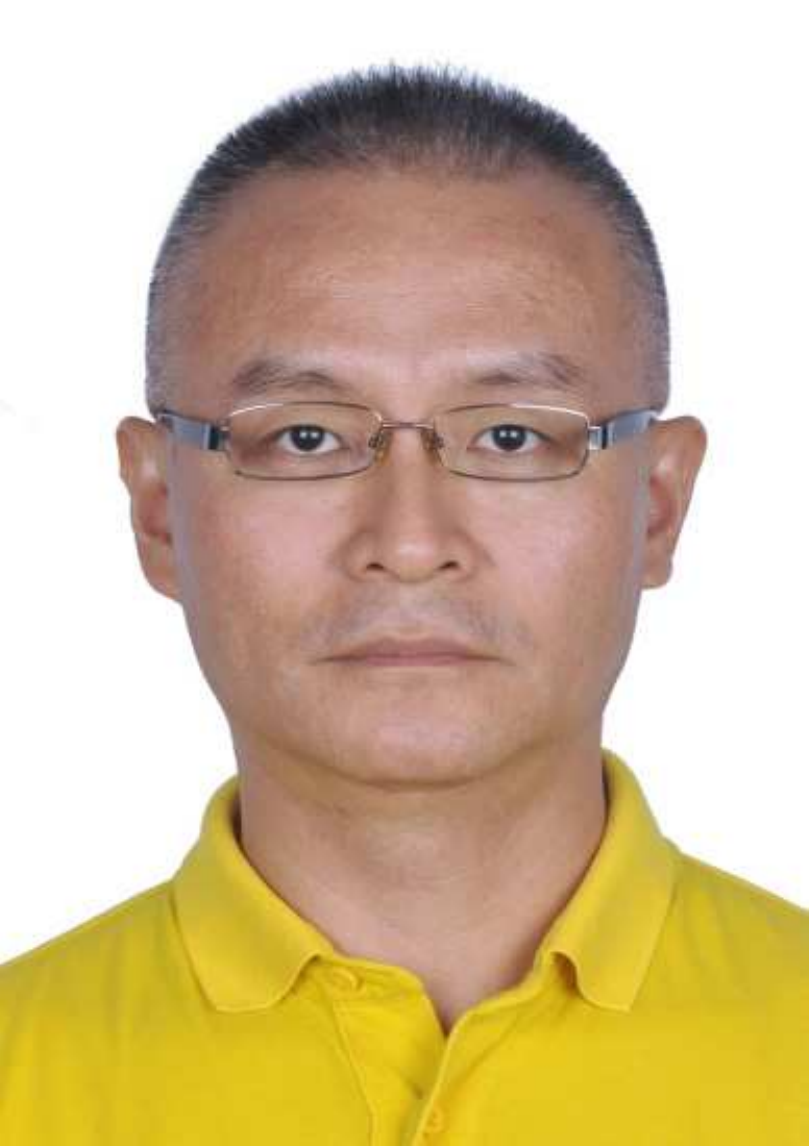}}]{Ming Tang}
(Member, IEEE) received the B.S.
degree in computer science and engineering and the
M.S. degree in artificial intelligence from Zhejiang
University, Hangzhou, China, in 1984 and 1987,
respectively, and the Ph.D. degree in pattern recognition and intelligence systems from the Chinese
Academy of Sciences, Beijing, China, in 2002.

He is currently a Professor with the Institute of
Automation, Chinese Academy of Sciences. His
current research interests include computer vision
and machine learning.
\end{IEEEbiography}

\begin{IEEEbiography}
[{\includegraphics[width=1in,height=1.25in,clip,keepaspectratio]{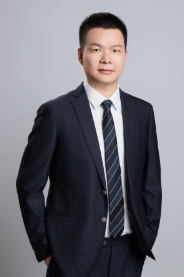}}]{Peng Su}
received his Ph.D. in Electronic and Information Engineering from The Chinese University of Hong Kong in 2020, specializing in computer vision and deep learning. He is currently Director of Autonomous Driving Research and Development at Huawei, leading the development of Physical AI systems for L4 autonomous driving.

\end{IEEEbiography}

\begin{IEEEbiography}
[{\includegraphics[width=1in,height=1.25in,clip,keepaspectratio]{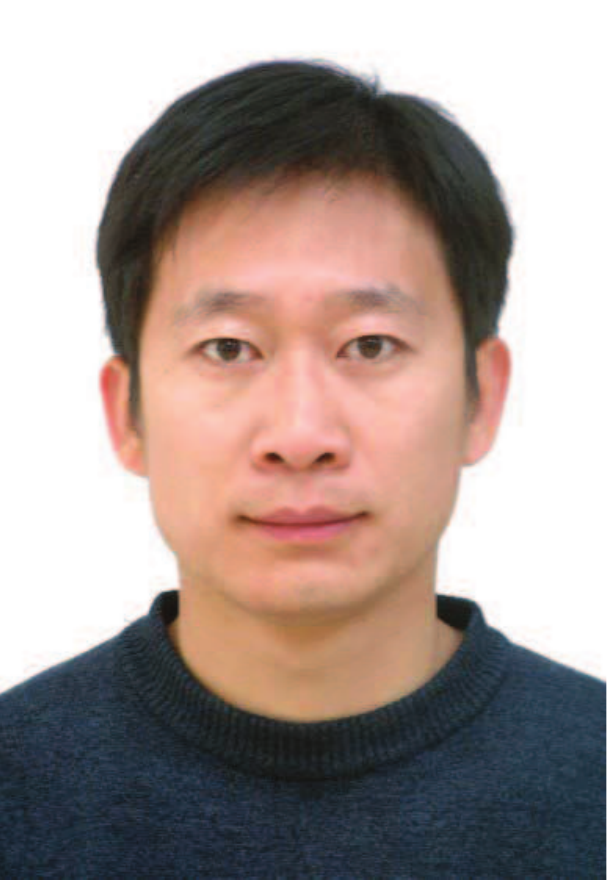}}]{Jinqiao Wang}
(Member, IEEE) received the B.E.
degree from Hebei University of Technology,
Tianjin, China, in 2001, the M.S. degree from Tianjin University, Tianjin, in 2004, and the Ph.D. degree
in pattern recognition and intelligence systems from
the National Laboratory of Pattern Recognition,
Chinese Academy of Sciences, Beijing, in 2008.

He is currently a Professor with the Chinese
Academy of Sciences. His research interests include
pattern recognition and machine learning, large multimodal models, and image and video analysis.
\end{IEEEbiography}


\vfill

\end{document}